\documentclass[10pt,twocolumn,letterpaper]{article}

\usepackage{cvpr}
\usepackage{times}
\usepackage{epsfig}
\usepackage{graphicx}

\usepackage{amssymb}
\usepackage{amsmath}
\usepackage{subcaption}
\usepackage{enumitem}

\usepackage{lipsum}
\usepackage{epigraph}
\usepackage{multirow}
\usepackage{booktabs}
\usepackage{tabularx}
\usepackage{colortbl}
\usepackage{epstopdf}
\usepackage{balance}
\usepackage[font=small,labelfont=bf]{caption}
\usepackage{lib}
\usepackage{url}

\AtBeginDocument{
  \addtolength{\abovedisplayskip}{-2ex}
  \addtolength{\abovedisplayshortskip}{-2ex}
  \addtolength{\belowdisplayskip}{-2ex}
  \addtolength{\belowdisplayshortskip}{-2ex}
}

\usepackage[pagebackref=true,breaklinks=true,letterpaper=true,colorlinks,bookmarks=false]{hyperref}
\setlist[enumerate]{leftmargin=*}
\newcommand{\vertical}[1]{\rotatebox[origin=l]{90}{\parbox{1.2cm}{#1}}}
\newcommand{\verticala}[1]{\rotatebox[origin=l]{90}{\parbox{1.2cm}{#1}}}

\newcommand{\rgbd}[0]{{\small RGB-D}\xspace}
\newcommand{\rgb}[0]{{\small RGB}\xspace}
\newcommand{\mcg}[0]{{\small MCG}\xspace}
\newcommand{\dd}[0]{{\small D}\xspace}

\newcommand{\nyu}[0]{{\small NYUD2}\xspace}
\newcommand{\jhmdb}[0]{{\small JHMDB}\xspace}
\newcommand{\ucf}[0]{{\small UCF 101}\xspace}
\newcommand{\alexnet}[0]{{\small AlexNet}\xspace}
\newcommand{\imagenet}[0]{{ImageNet}\xspace}
\newcommand{\regionAP}[0]{$AP^r$\xspace}

\newcommand{\boxAP}[0]{AP\xspace}
\newcommand{\vgg}[0]{{\small VGG}\xspace}
\newcommand{\rcnn}[0]{{\small R-CNN}\xspace}
\newcommand{\expt}[1]{{Exp. #1}\xspace}
\newcommand{\expta}[1]{{#1}\xspace}
\newcommand{\fastrcnn}[0]{{\small Fast R-CNN}\xspace}
\newcommand{\cnn}[0]{{\small CNN}\xspace}

\newcommand{\trainval}[0]{\textit{trainval}\xspace}

\newcommand{\test}[0]{\textit{test}\xspace}
\newcommand{\val}[0]{\textit{val}\xspace}
\newcommand{\hha}[0]{{\small HHA}\xspace}

\renewcommand{\ll}[1]{\texttt{\small #1}}

\newcommand{\supt}[0]{\textit{supervision transfer}\xspace}
\newcommand{\Supt}[0]{\textit{Supervision transfer}\xspace}
\newcommand{\tb}[1]{\textbf{#1}}
\newcommand{\pparagraph}[1]{\textbf{#1}}

\newcommand{\mscoco}[0]{{\small MS COCO}\xspace}
\newcommand{\insertW}[2]{\IfFileExists{#2}{\includegraphics[width=#1\textwidth]{#2}}{\includegraphics[width=#1\textwidth]{figures/blank.png}}}
\newcommand{\insertH}[2]{\IfFileExists{#2}{\includegraphics[height=#1\textwidth]{#2}}{\includegraphics[height=#1\textwidth]{figures/blank.png}}}
\newcommand{\insertHW}[3]{\IfFileExists{#2}{\includegraphics[height=#1\textwidth,width=#2\textwidth]{#3}}{\includegraphics[height=#1\textwidth,width=#2\textwidth]{figures/blank.png}}}
\newcommand{\insertFIG}[2]{\begin{figure*}\centering
\includegraphics[width=#2\linewidth]{vis/sds_test/#1.jpeg} \caption{Sample
detections and segmentation masks for #1 on \nyu \test set.} \figlabel{#1} \end{figure*}}

\cvprfinalcopy 

\begin{document}

\title{Cross Modal Distillation for Supervision Transfer}

\author{
Saurabh Gupta $\qquad$ Judy Hoffman $\qquad$ Jitendra Malik \\
University of California, Berkeley  \\
\texttt{\{sgupta, jhoffman, malik\}@eecs.berkeley.edu}
}

\maketitle

\begin{abstract}
In this work we propose a technique that transfers supervision between images
from different modalities. We use learned representations from a large labeled
modality as a supervisory signal for training representations for a new
unlabeled paired modality.  Our method enables learning of rich representations
for unlabeled modalities and can be used as a pre-training procedure for new
  modalities with limited labeled data.  We show experimental results where we
  transfer supervision from labeled \rgb images to unlabeled depth and optical
  flow images and demonstrate large improvements for both these cross modal
  supervision transfers. Code, data and pretrained models are available at
  \url{https://github.com/s-gupta/fast-rcnn/tree/distillation}.
\end{abstract}

\definecolor{Gray}{gray}{0.85}
\newcolumntype{g}{>{\columncolor{Gray}}c}

\section{Introduction}
Current paradigms for recognition in computer vision involve learning a generic
feature representation on a large dataset of labeled images, and then
specializing or finetuning the learned generic feature representation for the
specific task at hand. Successful examples of this paradigm include almost all
state-of-the-art systems: object detection \cite{girshickCVPR14}, semantic
segmentation \cite{long_shelhamer_fcn}, object segmentation
\cite{hariharanECCV14}, and pose estimation \cite{tulsianiCVPR15}, which start
from generic features that are learned on the ImageNet dataset
\cite{imagenet_cvpr09} using over a million labeled images and specialize them
for each of the different tasks.  Several different architectures for learning
these generic feature representations have been proposed over the years
\cite{krizhevsky2012imagenet,simonyan2014very,chatfield2014return}, but all
of these rely on the availability of a large dataset of labeled images to
learn feature representations.

The question we ask in this work is, what is the analogue of this paradigm for
images from modalities which do not have such large amounts of labeled data?
There are a large number of image modalities beyond \rgb images which are
dominant in computer vision, for example depth images coming from a Microsoft
Kinect, infra-red images from thermal sensors, aerial images from satellites and
drones, LIDAR point clouds from laser scanners, 
or even images of intermediate representations output from current
vision systems \eg optical flow and stereo images. The number of labeled
images from such modalities are at least a few orders of magnitude smaller than
the \rgb image datasets used for learning features, which raises the question: do
we need similar large scale annotation efforts to learn generic features for
images from each such different modality?

\setlength{\belowcaptionskip}{-14pt}
\begin{figure} \centering
\includegraphics[width=1.0\linewidth]{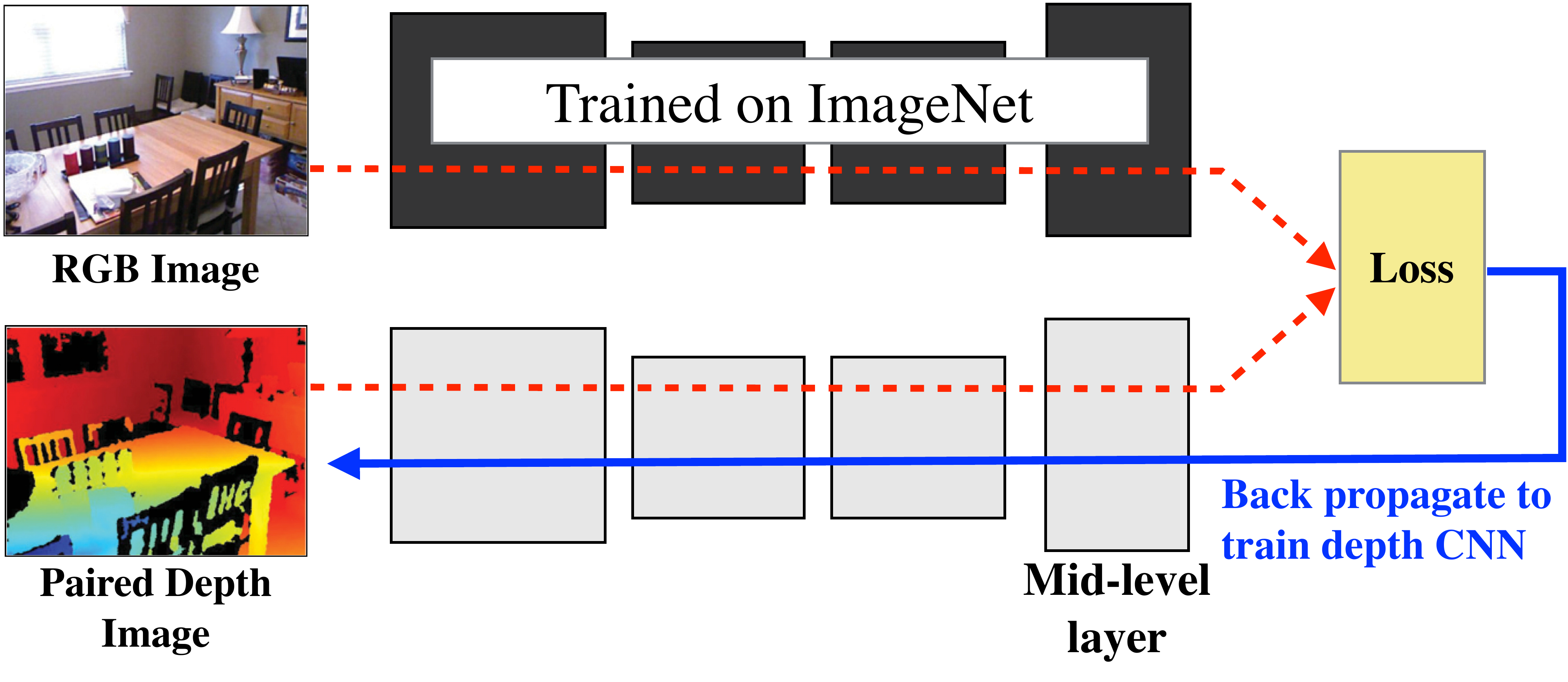}
\caption{\textbf{Architecture for supervision transfer}: We train a \cnn
model for a new image modality (like depth images), by teaching the network to
reproduce the mid-level semantic representations learned from a well labeled
image modality (such as \rgb images) for modalities for which there are paired
images.}
\figlabel{overview}
\end{figure}

We answer this question in this paper and propose a technique to transfer
learned representations from one modality to another. Our technique uses
`paired' images from the two modalities and utilizes the mid-level
representations from the labeled modality to supervise learning representations
on the paired un-labeled modality. We call our scheme
\textbf{\textit{supervision transfer}} and show that our learned
representations perform well on standard tasks like object detection. We also
show that our technique leads to learning useful feature hierarchies in the
unlabeled modality, which can be improved further with finetuning, and are
still complementary to representations in the source modality.  

As a motivating example, consider the case of depth images. While the largest
labeled \rgb dataset, ImageNet \cite{imagenet_cvpr09} consists of over a million
labeled images, the size of most existing labeled depth datasets is of the
order of a few thousands \cite{silbermanECCV14,song2015sun,b3do}. At the same
time there are a large number of unlabeled \rgb and depth image pairs. Our
technique leverages this large set of unlabeled paired images to transfer the
ImageNet supervision on \rgb images to depth images. Our technique is
illustrated in \figref{overview}.  We use a convolutional neural network that
has been trained on labeled images in the ImageNet dataset
\cite{imagenet_cvpr09}, and use the mid-level representation learned by these
{\cnn}s as a supervisory signal to train a \cnn on depth images. Our technique
for transferring supervision results in improvements in performance for the end
task of object detection on the \nyu dataset, where we improve the
state-of-the-art from 34.2\% to 41.7\% when using just the depth image and
from 46.2\% to 49.1\% when using both \rgb and depth images together.  We
report similar improvements for the task of simultaneous detection and
segmentation \cite{hariharanECCV14} and also show how supervision transfer
can be used for a zero-shot transfer of object detectors trained on \rgb
images to detectors that can run on depth images. 

Though we show detailed experimental results for supervision transfer from \rgb
to depth images, our technique is equally applicable to images from other
paired modalities. To demonstrate this, we show additional transfer results
from \rgb images to optical flow images where we improve mean average precision
for action detection on the \jhmdb dataset \cite{Jhuang:ICCV:2013} from 31.7\%
to 35.7\% when using just the optical flow image and no supervised
pre-training.

Our technique is reminiscent of the distillation idea from Hinton \etal
\cite{hinton2014distilling} (and its recent FitNets extension
\cite{romero2014fitnets}). Hinton \etal \cite{hinton2014distilling} extended
the model compression idea from Bucilua \etal \cite{bucilua2006model}
to what they call `distillation' and showed how large models trained on large
labeled datasets can be compressed by using the soft outputs from the large
model as targets for a much smaller model operating on the same modality. Our
work here is a generalization of this idea, and a) allows for transfer of
supervision at arbitrary semantic levels, and b) additionally enables transfer
of supervision between \textit{different} modalities using paired images. More
importantly, our work here allows us to extend the success of recent deep \cnn
architectures to new imaging modalities without having to collect large scale
labeled datasets necessary for training deep {\cnn}s.

\section{Related Work}

There has been a large body of work on transferring knowledge between different
visual domains, belonging to the \textit{same} modality. Initial work
\cite{kulis-cvpr11, gong-cvpr12, aytar-iccv11, duan-icml12, hoffman-iclr13}
studied the problem in context of shallow image representations. While
\cite{kulis-cvpr11, gong-cvpr12} sought to learn transformations between well
labeled source and sparsely labeled target domains, Aytar \etal \cite{aytar-iccv11} 
use the source models as a parameter regularizer for target models, \cite{duan-icml12,
hoffman-iclr13} combine these two approaches into a single joint optimization
problem. Chopra \etal~\cite{ref:dlid} introduced one of the first deep
architectures for visual adaptation by replicating feature extraction for each
domain and producing intermediate interpolated domains, while Ghifary
\etal~\cite{da-mmd} showed a single layer neural net could be used to learn the
feature transformation between simple domain shifts.

More recently, with the introduction of supervised CNN models by Krizhevsky
\etal~\cite{krizhevsky2012imagenet}, the community has been moving towards a
generic set of features which can be specialized to specific tasks and domains
at hand \cite{donahueDecaf, girshickCVPR14, fastrcnn, overfeat, lsda} and
traditional visual adaptation techniques can be used in conjunction with such
features \cite{hoffman-iclr14}.  More recently, unsupervised domain adaptation
techniques have been introduced which learn to adapt deep representations so as
to minimize the discrepancy between the source and target
distributions~\cite{ddc, ganin2015, long2015dan}. 

All these lines of work study and solve the problem of domain adaptation within
the same modality. In contrast, our work here tackles the problem of domain
adaptation across \textit{different} modalities. Most methods for
intra-modality domain adaptation described above start from an initial set of
features on the target domain, and \apriori it is unclear how this can be done
when moving across modalities, limiting the applicability of aforementioned
approaches to our problem. This cross-model transfer problem has received much
less attention. Notable among those include \cite{christoudias2010learning,
ngiam2011multimodal, JMLR:v15:srivastava14b, socher2013zero, frome2013devise}.
While \cite{christoudias2010learning, JMLR:v15:srivastava14b} hallucinate
modalities during training time, \cite{ngiam2011multimodal, socher2013zero,
frome2013devise} focus on the problem of jointly embedding or learning
representations from multiple modalities into a shared feature space to improve
learning \cite{ngiam2011multimodal} or enabling zero-shot
learning\cite{socher2013zero, frome2013devise}. Our work here instead transfers
high quality representations learned from a large set of labeled images of one
modality to completely unlabeled images from a new modality, thus leading to a
generic feature representations on the new modalities which we show are useful
for a variety of tasks.

\section{Supervision Transfer}
\seclabel{method}
Let us assume we have a modality $\mathcal{M}_d$ with unlabeled data, $D_d$ for
which we would like to train a rich representation.  We will do so by
transferring information from a separate modality, $\mathcal{M}_s$, which has a
large labeled set of images, $\mathcal{D}_s$, and a corresponding $K$ layered
rich representation. We assume this rich representation is layered although our
proposed method will work equally well for non-layered representations. We use
convolutional neural networks as our layered rich representation.

We denote this image representation as $\Phi = \{\phi^{i}_{\mathcal{M}_s,D_s} i
\in [1 \ldots K]\}$.  $\phi^{i}_{\mathcal{M}_s,D_s}$ is the $i^{th}$ layer
representation for modality $\mathcal{M}_s$ which has been trained on labeled
images from dataset $D_s$, and it maps an input image from modality
$\mathcal{M}_s$ to a feature vector in $\mathbb{R}^{n_i}$.
\begin{eqnarray}
\phi^i_{\mathcal{M}_s,D_s} \; {:} \; \mathcal{M}_s \mapsto \mathbb{R}^{n_i} 
\end{eqnarray}

Feature vectors from consecutive layers in such layered representations are
related to one another by simple operations like non-linearities, convolutions,
pooling, normalizations and dot products (for example layer 2 features may be
related to layer 1 features using a simple non-linearity like $\max$ with $0$:
$\phi^{2}_{\mathcal{M}_s,D_s}(x) = \max(0, \phi^{1}_{\mathcal{M}_s,D_s}(x))$).  Some
of these operations like convolutions and dot products have free parameters.
We denote such parameters associated with operation at layer $i$ by ${w^i_s}$.
The structure of such architectures (the sequence of operations, and the size
of representations at each layer, \etc) is hand designed or validated using
performance on an end task. Such validation can be done on a small set of
annotated images. Estimating the model parameters ${w^i_s}$ is much more
difficult. The number of these parameters for most reasonable image models can
easily go up to a few millions. Heretofore, state-of-the-art models require
discriminative learning of these parameters using a large labeled training set. 

Now suppose we want to learn a rich representation for images from modality
$\mathcal{M}_d$, for which we do not have access to a large dataset of labeled
images. We assume we have already hand designed an appropriate architecture
$\Psi = \{\psi^{i}_{\mathcal{M}_d} \; \forall i \in [1 \ldots L]\}$.  The task then is to
effectively learn the parameters associated with various operations in the
architecture, without having access to a large set of annotated images for
modality $\mathcal{M}_d$. As before, we denote these parameters to be learned by $W^{[1
\ldots L]}_d = \{w^i_d \; \forall i \in [1 \ldots L]\}$

In addition to $D_s$, let us assume that we have access to a large dataset of
\emph{un-annotated paired} images from modalities $\mathcal{M}_s$ and
$\mathcal{M}_d$. We denote this dataset by $U_{s,d}$. By paired images we mean
a set of images of the same scene in two different modalities. Our proposed
scheme for training rich representations for images of modality $\mathcal{M}_d$
is to learn the representation $\Psi$ such that the image representation
$\psi^{L}_{\mathcal{D}}(I_d)$ for image $I_d$ matches the image representation
$\phi^{i^*}_{\mathcal{M}_s,D_s}(I_s)$ for its image pair $I_s$ in modality
$\mathcal{M}_s$ for some chosen and fixed layer $i^* \in [1 \ldots K]$. We
measure the similarity between the representations using an appropriate
loss function $f$ (for example, euclidean loss). Note that the
representations $\psi^{i^*}_{\mathcal{M}_s}$ and $\phi^{L}_{\mathcal{M}_d}$ may
not have the same dimensions. In such cases we embed features
$\psi^{L}_{\mathcal{M}_d}$ into a space with the same dimension as
$\phi^{i^*}_{\mathcal{M}_s}$ using an appropriate simple transformation
function $t$ (for example a linear or affine function).
\begin{eqnarray}
\min_{W^{[1 \ldots L]}_d} \sum_{(I_s, I_d) \in
{U_{s,d}}}{f\left( t \left(\psi^{L}_{\mathcal{M}_d}(I_d) \right), \;
\phi^{i^*}_{\mathcal{M}_s,D_s}(I_s)\right)}
\end{eqnarray}
We call this process \emph{supervision transfer} from layer $i^*$ in $\Phi$
of modality $\mathcal{M}_s$ to layer $L$ in $\Psi$ of modality
$\mathcal{M}_d$.  

The recent distillation method from Hinton \etal \cite{hinton2014distilling} is
a specific instantiation of this general method, where a) they focus on the
specific case when the two modalities $\mathcal{M}_s$ and $\mathcal{M}_d$ are
the same and b) the \emph{supervision transfer} happens at the very last
prediction layer, instead of an arbitrary internal layer in representation
$\Phi$.

Our experiments in \secref{experiments} demonstrate that this proposed method
for transfer of supervision is a) effective at learning good feature
hierarchies, b) these hierarchies can be improved further with finetuning,
and c) the resulting representation can be complementary to the
representation in the source modality $\mathcal{M}_s$ if the modalities
permit.

\section{Experiments}
\seclabel{experiments}
In this section we present experimental results for
the \nyu dataset where we use color and depth images as the paired modalities,
and on the \jhmdb video dataset for which we use the \rgb and optical flow
frames as the two modalities.

\renewcommand{\arraystretch}{1.5} 
\begin{table*}
\centering
\resizebox{1.0\linewidth}{!}{
\begin{tabular}{l>{\raggedright}p{0.24\textwidth}>{}cl l>{\raggedright}p{0.28\textwidth}>{}cl l>{\raggedright}p{0.24\textwidth}>{}cl}
  \toprule
  \multicolumn{3}{c}{Does supervision transfer work?} & & 
  \multicolumn{3}{c}{How good is the transferred representation by itself?} & & 
  \multicolumn{3}{c}{Are the representations complementary?} \\
  \cmidrule{1-3} \cmidrule{5-7} \cmidrule{9-11}
  \expt{1A}   & no init                                                                    & 22.7    &  &
  \expt{2A}   & copy from \rgb (ft \ll{fc} only)                                           & 19.8    &  &
  \expt{3A}   & [RGB]: \rgb network on \rgb images \alexnet                                & 22.3 \\
  \expt{1B}   & copy from \rgb                                                             & 25.1    &  &
  \expt{2B}   & supervision transfer (ft \ll{fc} only) \alexnet$^*$~$\rightarrow$~\alexnet & 30.0    &  &
  \expt{3B}   & [RGB] + copy from \rgb                                                     & 33.8 \\
  \expt{1C}   & supervision transfer $\qquad$ \alexnet~$\rightarrow$~\alexnet              & 29.7    &  &
  \expt{2C}   & supervision transfer (ft \ll{fc} only) \vgg$^*$~$\rightarrow$~\alexnet     & 32.2    &  &
  \expt{3C}   & [RGB] + supervision transfer \alexnet$^*$~$\rightarrow$~\alexnet           & 35.6 \\
  \expt{1D}   & supervision transfer \alexnet~$^*$~$\rightarrow$~\alexnet                  & 30.5    &  &
  \expt{2D}   & supervision transfer \vgg~$^*$~$\rightarrow$~\alexnet                      & 33.6    &  &
  \expt{3D}   & [RGB]+ supervision transfer \vgg~$^*$~$\rightarrow$~\alexnet               & 37.0 \\
  \bottomrule
\end{tabular}}
\caption{\small We evaluate different aspects of our \textit{supervision
transfer} scheme on the object detection task on the \nyu \val set using the
mAP metric. Left column demonstrates that our scheme for pre-training is better
than alternatives like no pre-training, and copying over weights from \rgb
networks. The middle column demonstrates that our technique leads to transfer
of mid-level semantic features which by themselves are highly discriminative,
and that improving the quality of the supervisory network translated to
improvements in the learned features. Finally, the right column demonstrates
that the learned features on the depth images are still complementary to the
features on the \rgb image they were supervised with.} 
\tablelabel{control}
\end{table*}

Our general experimental framework consists of two steps. The first step is
\textit{supervision transfer} as proposed in \secref{method}, and the second
step is to assess the quality of the transferred representation by using it for
a downstream task. For both of the datasets we study, we consider the domain of
\rgb images as $\mathcal{M}_s$ for which there is a large dataset of labeled
images $D_s$ in the form of ImageNet \cite{imagenet_cvpr09}, and treat depth
and optical flow respectively as $\mathcal{M}_d$. These choices for
$\mathcal{M}_s$ and $\mathcal{M}_d$ are of particular practical significance,
given the lack of large labeled datasets for depth images and optical flow, at
the same time, the abundant availability of paired images coming from \rgbd
sensors (for example Microsoft Kinect) and videos on the Internet respectively. 

For our layered image representation models, we use convolutional neural
networks ({\cnn}s) \cite{lecun-89e,krizhevsky2012imagenet}. These networks have
been shown to be very effective for a variety of image understanding tasks
\cite{donahueDecaf}. We experiment with the network architectures from
Krizhevsky \etal \cite{krizhevsky2012imagenet} (denoted \alexnet), Simonyan and
Zisserman \cite{simonyan2014very} (denoted \vgg), and use the models
pre-trained on ImageNet \cite{imagenet_cvpr09} from the Caffe \cite{jiaCaffe}
Model Zoo.

We use an architecture similar to \cite{krizhevsky2012imagenet} for the layered
representations for depth and flow images. We do this in order to be able to
compare to past works which learn features on depth and flow images
\cite{guptaECCV14,gkioxari2015finding}. Validating different \cnn architectures
for depth and flow images is a worthwhile scientific endeavor, which has not
been undertaken so far, primarily because of lack of large scale labeled
datasets for these modalities. Our work here provides a method to circumvent
the need for a large labeled dataset for these and other image modalities,
and will naturally enable exploring this question in the future, however we do not
delve in this question in the current work.

We next describe our design choices for which layers to transfer supervision
between, and the specification of the loss function $f$ and the transformation
function $t$. We experimented with what layer to use for transferring
supervision, and found transer at mid-level layers works best, and use the last
convolutional layer \ll{pool5} for all experiments in the paper. Such a choice
also resonates well with observations from \cite{lenc15understanding} that
lower layers in {\cnn}s are modality specific (and thus harder to transfer
across modalities) and visualizations from \cite{girshickCVPR14} that neurons
in mid-level layers are semantic and respond to parts of objects. Transferring
at \ll{pool5} also has the computational benefit that training can be
efficiently done in a fully convolutional manner over the whole image.

For the function $f$, we use $L2$ distance between the feature vectors,
$f(\mathbf{x}, \mathbf{y}) = \|\mathbf{x} - \mathbf{y}\|^2_2$. We also
tried $f(\mathbf{x}, \mathbf{y}) = \mathbf{1}(\mathbf{y} >
\tau) \cdot \log p(\mathbf{x}) + \mathbf{1}(\mathbf{y} \le
\tau) \cdot \log(1-p(\mathbf{x}))$ (where $p(x) = \frac{e^{\alpha x}}{1+e^{\alpha
x}}$, $\mathbf{1}(x)$ is the indicator function), for some reasonable choices
of $\alpha$ and $\tau$ but that resulted in worse performance in initial
  experiments and we did not experiment with it further.

Finally, the choice of the function $t$ varies with different pairs of
networks. As noted above, we train using a fully convolutional architecture.
This requires the spatial resolution of the two layers $i^*$ in $\Phi$ and $L$
in $\Psi$ to be similar, which is trivially true if the architectures $\Phi$
and $\Psi$ are the same. When they are not (for example when we transfer from
\vgg net to \alexnet), we adjust the padding in the \alexnet to obtain the same
spatial resolution at \ll{pool5} layer. 

This apart, we introduce an adaptation layer comprising of $1 \times 1$
convolutions followed by $ReLU$ to map from the representation at layer $L$ in
$\Psi$ to layer $i^*$ in $\Phi$. This accounts for difference in the number of
neurons (for example when adapting from \vgg to \alexnet), or even when the
number of neurons are the same, allows for domain specific fitting. For \vgg to
\alexnet transfer we also needed to introduce a scaling layer to make the
average norm of features comparable between the two networks.

\setlength{\belowcaptionskip}{-14pt}
\begin{figure*}
\centering
\renewcommand{\arraystretch}{1} 
\setlength{\tabcolsep}{2pt}
{\scriptsize
\begin{tabular}{cccccccc}
\includegraphics[width=0.154\linewidth]{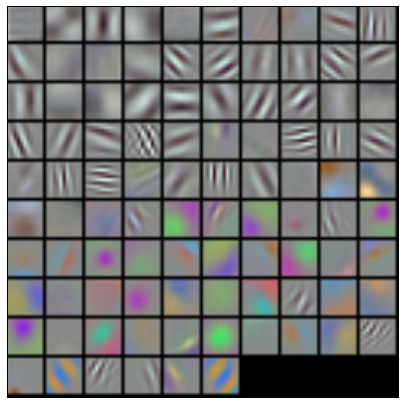} &
\includegraphics[width=0.154\linewidth]{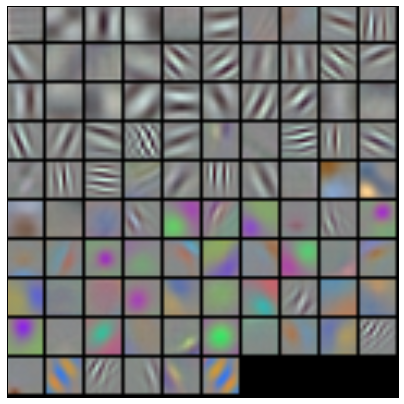} &
\includegraphics[width=0.154\linewidth]{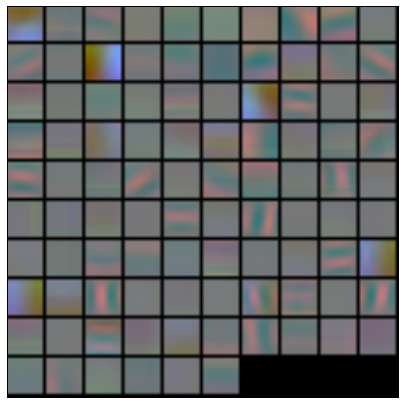} &
\includegraphics[width=0.154\linewidth]{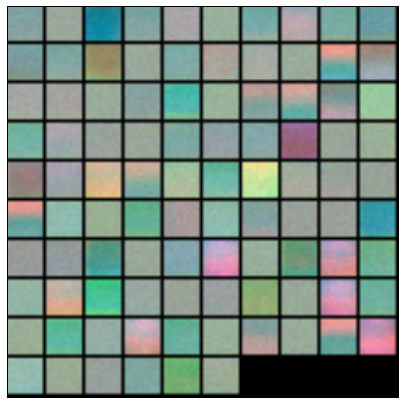} & & &
\includegraphics[width=0.154\linewidth]{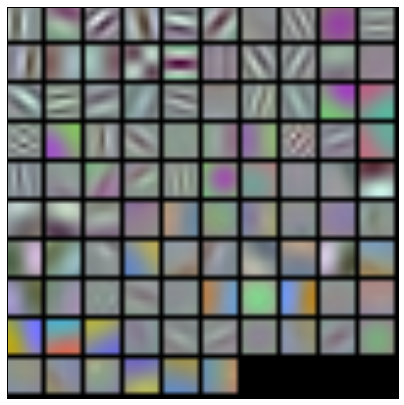} &
\includegraphics[width=0.154\linewidth]{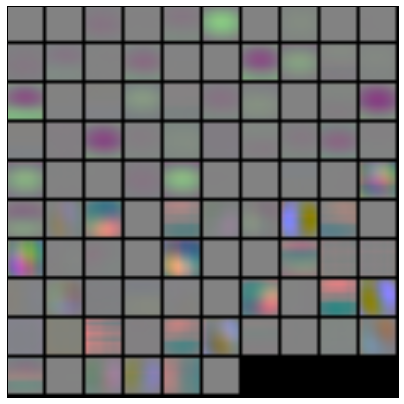} \\
(a) & (b) & (c) & (d) & & & (e) & (f) \\
\end{tabular}
\begin{tabular}{ccccc}
\includegraphics[width=0.315\linewidth]{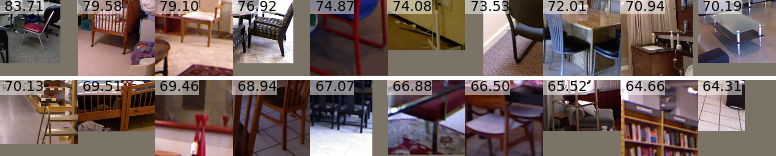} & & &
\includegraphics[width=0.315\linewidth]{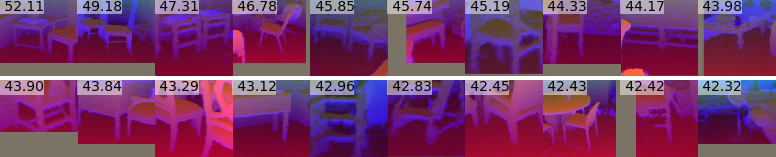} &
\includegraphics[width=0.315\linewidth]{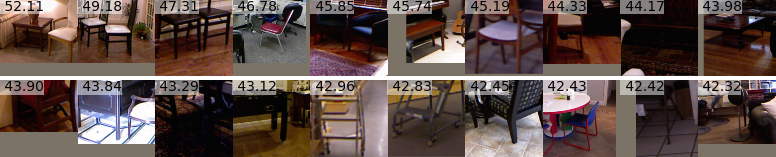} \\
\\
\includegraphics[width=0.315\linewidth]{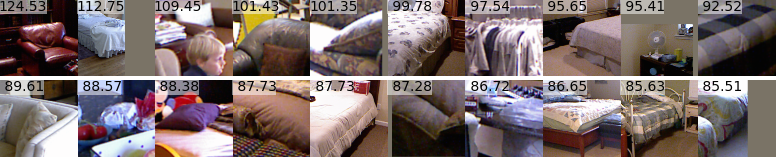} & & &
\includegraphics[width=0.315\linewidth]{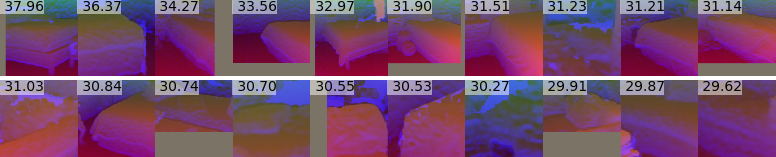} &
\includegraphics[width=0.315\linewidth]{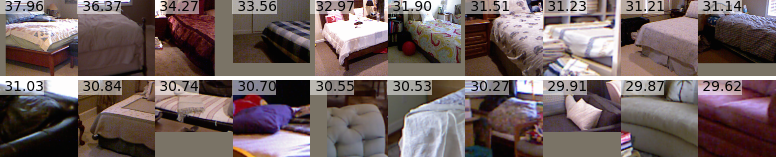} \\
(g) & & & (h) & (i) \end{tabular}} \caption{\small \textbf{Visualization of
learned filters} (best viewed in color): (a) visualizes filters learned on \rgb
images from ImageNet data by \alexnet. (b) shows these filters after the
finetuning on \hha images, and hardly anything changes visually. (c) shows \hha
image filters from our pre-training scheme, which are much different from ones
that are learned on \rgb images. (d) shows \hha image filters learned without
any pre-training. (e) shows optical flow filters learned by
\cite{gkioxari2015finding}. Note that they initialize these filters from \rgb
filters and these also do not change much over their initial \rgb filters. (f)
shows filters we learn on optical flow images, which are again very different
from filters learned on \rgb or \hha images. (g) shows image patches
corresponding to highest scoring activations for two neurons in the {\rgb}
{\cnn}. (h) shows \hha image patches corresponding to highest scoring
activations of the same neuron in the supervision transfer depth {\cnn}. (i)
shows the corresponding \rgb image patch for these depth image patches for ease
of visualization.}
\figlabel{filters}
\end{figure*}

\subsection{Transfer to Depth Images}
\seclabel{nyu}
We first demonstrate how we transfer supervision from color images to depth
images as obtained from a range sensor like the Microsoft Kinect. As described
above, we do this set of experiments on the \nyu dataset \cite{silbermanECCV12}
and show results on the task of object detection and instance segmentation
\cite{guptaECCV14}. The \nyu dataset consists of 1449 paired \rgb and \dd
images. These images come from 464 different scenes and were hand selected from
the full video sequence while ensuring ensure diverse scene content
\cite{silbermanECCV12}. The full video sequence that comes with the dataset has
over 400K \rgbd frames, we use 10K of these frame pairs for supervision transfer.

In all our experiments we report numbers on the standard \val and \test splits
that come with the dataset \cite{silbermanECCV12,guptaECCV14}. Images in these
splits have been selected while ensuring that all frames belonging to the same
scene are contained entirely in exactly one split. We additionally made sure
only frames from the corresponding training split were used for supervision
transfer.

The downstream task that we study here is that of object detection. We follow
the experimental setup from Gupta \etal \cite{guptaECCV14} for object detection
and study the 19 category object detection problem, and use mean average
precision (mAP) to measure performance.

\pparagraph{Baseline Detection Model} We use the model from Gupta \etal
\cite{guptaECCV14} for object detection.  Their method builds off \rcnn
\cite{girshickCVPR14}.  In our initial experiments we adapted their model to
the more recent \fastrcnn framework \cite{fastrcnn}. We summarize our key
findings here. First, \cite{guptaECCV14} trained the final detector on both
\rgb and \dd features jointly. We found training independent models all the way
and then simply averaging the class scores before the SoftMax performed better.
While this is counter-intuitive, we feel it is plausible given limited amount
of training data. Second, \cite{guptaECCV14} use features from the \ll{fc6}
layer and observed worse performance when using \ll{fc7} representation; in
our framework where we are training completely independent detectors for the
two modalities, using \ll{fc7} representation is better than using
\ll{fc6} representation.  Finally, using bounding box regression boosts
performance. Here we simply average the predicted regression target from the
detectors on the two modalities. All this analysis helped us boost the mean
\boxAP on the \test set from 38.80\% as reported by
\cite{guptaECCV14,guptaCVPR15a} to 44.39\%, using the same \cnn network and
supervision. This already is the state-of-the-art result on this dataset and we
use this as a baseline for the rest of our experiments. We denote this model as
`\cite{guptaECCV14} + \fastrcnn'. We followed the default setup for training
\fastrcnn, $40K$ iterations, base learning rate of $0.001$ and stepping it down
by a factor of $10$ after every $30K$ iterations, except that we finetune all
the layers, and use $688$px length for the shorter image side. We used \rgbd box
proposals from \cite{guptaECCV14} for all experiments.

Note that Gupta \etal \cite{guptaECCV14} embed depth images into a geocentric
embedding which they call \hha (\hha encodes horizontal disparity, height above
ground and angle with gravity) and use the \alexnet architecture to learn  \hha
features and \textit{copy over the weights from the \rgb~\cnn that was trained
for 1000 way classification \cite{krizhevsky2012imagenet} on ImageNet
  \cite{imagenet_cvpr09}} to initialize this network. All through this paper,
  we stick with using \hha embedding\footnote{We use the term depth and \hha
  interchangeably.} to represent the input depth images, and their network
  architecture, and show how our proposed \textit{supervision transfer} scheme
  improves performances over their technique for initialization. We summarize
  our various transfer experiments below:

\pparagraph{Does \textit{supervision transfer} work?}
The first question we investigate is if we are able to transfer supervision to
a new modality. To understand this we conducted the following three
experiments:
  
  \textbf{1. no init (\expta{1A})}: randomly initialize the depth network
  using weight distributions typically used for training on \imagenet and
  simply train this network for the final task. While training this network we
  train for 100K iterations, start with a learning rate on 0.01 and step it
  down by a factor of 10 every 30K iterations. 
  
  \textbf{2. copy from \rgb (\expta{1B})}: copy weights from a \rgb network
  that was trained on ImageNet. This is same as the scheme proposed in
  \cite{guptaECCV14}.  This network is then trained using the standard
  \fastrcnn settings.  
  
  \textbf{3. \textit{supervision transfer} (\expta{1C})}: train layers
  \ll{conv1} through \ll{pool5} from random initialization using the
  \textit{supervision transfer} scheme as proposed in \secref{method}, on the
  5K paired \rgb and \dd images from the video sequence from \nyu for scenes
  contained in the training set. We then plug in these trained layers 
  along with randomly initialized \ll{fc6}, \ll{fc7} and classifier
  layers for training with \fastrcnn.
We report the results in \tableref{control}. We see that `copy from \rgb'
surprisingly does better than `no init', which is consistent with what Gupta
\etal report in \cite{guptaECCV14}, but our scheme for \textit{supervision
transfer} outperforms both these baselines by a large margin pushing up mean AP
from 25.1\% to 29.7\%. We also experimented with using a \rgb network $\Psi$
that has been adapted for object detection on this dataset for supervising the
transfer (\expta{1D}) and found that this boosted performance further from
29.7\% to 30.5\% (\expta{1D} in \tableref{control}, $\mbox{\alexnet}^*$
indicates \rgb~\alexnet that has been adapted for
detection on the dataset). We use this scheme for all subsequent experiments.  

\pparagraph{Visualizations.} We visualize the filters from the first layer for
these different schemes of transfer in \figref{filters}(a-f), and observe that
our training scheme learns reasonable filters and find that these filters are
of different nature than filters learned on \rgb images. In contrast, note that
schemes which initialize depth {\cnn}s with {\rgb} {\cnn}s weights, filters in
the first layer change very little.  We also visualize patches giving high
activations for neurons paired across \rgb and \dd images
\figref{filters}(g-i). High scoring patches from {\rgb} {\cnn} (\alexnet in this
case), correspond to parts of object (g), high scoring patches from the depth
\cnn also corresponds to parts of the same object class (h and i). 

\pparagraph{How good is the transferred representation by itself?} The next
question we ask is if our \textit{supervision transfer} scheme 
transfers good representations or does it only provide a good initialization
for feature learning. To answer this question, we conducted the following 
experiments:
  
  \textbf{1. Quality of transferred \ll{pool5} representation (\expta{2A,
  2B})}: The first experiment was to evaluate the quality of the transferred
  \ll{pool5} representation. To do this, we froze the network parameters for
  layers \ll{conv1} through \ll{pool5} to be those learned during the transfer
  process, and only learn parameters in \ll{fc6}, \ll{fc7} and classifier
  layers during \fastrcnn training (\expta{2B} `supervision transfer adapted
  (ft \ll{fc} only)'). We see that there is only a moderate degradation in
  performance for our learned features from 30.5\% (\expta{1D}) to 30.0\%
  (\expta{2B}) indicating that the features learned on depth images at
  \ll{pool5} are discriminative by themselves. In contrast, when freezing
  weights when copying from ImageNet (\expta{2A}), performance degrades
  significantly to 19.8\%. 
  
  \textbf{2. Improved transfer using better supervising network $\Phi$
  (\expta{2C, 2D})}: The second experiment investigated if performance improves
  as we improve the quality of the supervising network. To do this, we
  transferred supervision from \vgg net instead of \alexnet (\expta{2C})
  \footnote{To transfer from \vgg to \alexnet, we use $150K$ transfer
  iterations instead of $100K$.  Running longer helps for \vgg to \alexnet
  transfer by $1.5\%$ and much less (about 0.5\%) for \alexnet to \alexnet
  transfer.}. \vgg net has been shown to be better than \alexnet for a variety
  of vision tasks. As before we report performance when freezing parameters till
  \ll{pool5} (\expta{2C}), and learning all the way (\expta{2D}). We see that
  using a better supervising net results in learning better features for depth
  images: when the representation is frozen till \ll{pool5} we see performance
  improves from 30.0\% to 32.2\%, and when we finetune all the layers
  performance goes up to 33.6\% as compared to 30.5\% for \alexnet. 

\renewcommand{\arraystretch}{1.4} 
\setlength{\tabcolsep}{4pt}
\begin{table}
\centering
\footnotesize
\resizebox{1.0\linewidth}{!}{
\begin{tabular}{lcccc} \toprule
\val & \multicolumn{2}{c}{\regionAP at 0.5} & \multicolumn{2}{c}{\regionAP at 0.7} \\ \cmidrule(r){2-3}  \cmidrule(r){4-5}
& \ll{~~~~fc7~~~~} & \ll{+pool2+conv4} & \ll{~~~~fc7~~~~} & \ll{+pool2+conv4} \\
\midrule
\rgb & 26.3 & 29.8 & 14.8 & 18.3 \\
\dd  & 28.4 & 31.5 & 17.4 & 19.6 \\ \bottomrule
\end{tabular}}
\caption{\small \textbf{Region detection average precision \regionAP on {\nyu}
{\val} set}: Performance on {\nyu} {\val} set where we observe similar
boosts in performance when using hyper-column transform with our learned feature
hierarchies (learned using supervision transfer on depth images) as obtained
with more standard feature hierarchies learned on \imagenet on \rgb images.}
\tablelabel{sds-control}
\end{table}

\renewcommand{\arraystretch}{1.4} 
\setlength{\tabcolsep}{4pt}
\begin{table}
\centering
\footnotesize
\resizebox{1.0\linewidth}{!}{
\begin{tabular}{lccccc} \toprule
\test & modality & \rgb Arch. & \dd Arch. & \regionAP at 0.5 & \regionAP at 0.7 \\
\midrule
\cite{hariharan2015hypercolumns}    & \rgb       & \alexnet & -        & 23.4 & 13.4 \\
Gupta \etal~\cite{guptaCVPR15a}     & \rgb + \dd & \alexnet & \alexnet & 37.5 & 21.8 \\
Our (\textit{supervision transfer}) & \rgb + \dd & \alexnet & \alexnet & \tb{40.5} & \tb{25.4} \\
\midrule
\cite{hariharan2015hypercolumns}    & \rgb       & \vgg     & -        & 31.0 & 17.7 \\
Our (\textit{supervision transfer}) & \rgb + \dd & \vgg     & \alexnet & \tb{42.1}  & \tb{26.9}\\
\bottomrule
\end{tabular}}
\caption{\small \textbf{Region detection average precision \regionAP on \nyu \test set.}}
\tablelabel{sds-test-summary}
\end{table}

\pparagraph{Is the learned representation complementary to the representation
on the source modality?} The next question we ask is if the representation
learned on the depth images complementary to the representation on the \rgb
images from which it was learned. To answer this question we look at the
performance when using both the modalities together. We do this the same way
that we describe for the baseline model and simply average the category scores
and regression targets from the \rgb and \dd detectors.
\tableref{control}(right) reports our findings. Just using \rgb images (\expta{3A}) gives us
a performance of 22.3\%. Combining this with the \hha network as initialized
using the scheme from Gupta \etal \cite{guptaECCV14} (\expta{3B}) boosts performance to
33.8\%. Initializing the \hha network using our proposed supervision transfer
scheme when transferring from $\mbox{\alexnet}^*$ to \alexnet (\expta{3C}) gives us 35.6\% and when
transferring from $\mbox{\vgg}^*$ to \alexnet (\expta{3D}) gives us 37.0\%. These results show that
the representations are still complementary and using the two together can help
the final performance. 

\pparagraph{Does supervision transfer lead to meaningful intermediate layer
representations?} The next questions we investigate is if the intermediate
layers learned in the target domain $\mathcal{M}_d$ through \supt carry useful
information. \cite{koenderink1987representation} hypothesize that information
from intermediate layers in such hierarchies carry information which may be
useful for fine grained tasks. Jones and Malik \cite{jones1992determining} and
Weber and Malik \cite{weber1995robust} and in more recent work Hariharan \etal
\cite{hariharan2015hypercolumns} and Long \etal \cite{long_shelhamer_fcn}
operationalize this and demonstrate improvements for fine grained tasks like
correspondence estimation and segmentation. Here we investigate if the
representations learned using {\supt} also share this property. To test this,
we follow the hyper-column architecture from Hariharan \etal
\cite{hariharan2015hypercolumns} and study the task of simultaneous detection
and segmentation (SDS) \cite{hariharanECCV14} and investigate if the use of
hyper-columns with our trained networks results in similar improvements as
obtained when using more traditionally trained {\cnn}s. We report the results
in \tableref{sds-control}. On the \nyu dataset, the hyper-column transform
improves \regionAP from 26.3\% to 29.8\% when using \alexnet for \rgb images.
We follow the same experimental setup as proposed in
\cite{Hariharan:EECS-2015-193}, and fix the \cnn parameters (to a network
that was finetuned for detection on \nyu dataset) and only learn the classifier
parameters and use features from \ll{pool2} and \ll{conv4} layers in addition
to \ll{fc7} for figure ground prediction. When doing the same for our \supt
network we observe a similar boost in performance from 28.4\% to 31.5\% when
using the hyper-column transform. This indicates that models trained using \supt
not only learn good representations at the point of supervision transfer
(\ll{pool5} in this case), but also in the intermediate layers of the network.

\renewcommand{\arraystretch}{1.4} 
\setlength{\tabcolsep}{4pt}
\begin{table}
\centering
\footnotesize
\resizebox{\linewidth}{!}{
\begin{tabular}{cccccccc} \toprule
\ll{pool1} & \ll{pool2}  & \ll{conv3} & \ll{conv4} & \ll{pool5} & \ll{~fc6~} &
\ll{~fc7~} & \ll{conv3} + \ll{fc7} \\
24.4 & 28.4 & 30.6 & 29.9 & 30.5 & 29.7 & 27.7 & 31.3 \\ \bottomrule
\end{tabular}}
\caption{\small \textbf{Mean \boxAP on \nyu \val set as a function of layer used for
supervision transfer.}}
\tablelabel{layer-sweep}
\end{table}

\pparagraph{How does performance vary as the transfer point is changed?} We now
study how performance varies as we vary the layer used for supervision
transfer. We stick to the same experimental setup as used for \expt{1D} in
\tableref{control}, and conduct supervision transfer at different layers of the
network.  Layers above the transfer point are initialized randomly and learned
during detector training. For transferring features from layers 1 to 5, we use
fully convolutional training as before. But when transferring \ll{fc6} and
\ll{fc7} features we compute them over bounding box proposals (we use
\rgbd~\mcg bounding box proposals \cite{guptaECCV14}) using Spatial Pyramid
Pooling on \ll{conv5} \cite{kaiming2014spatial,fastrcnn}.

We report the obtained \boxAP on the \nyu \val set in \tableref{layer-sweep}.
We see performance is poor when transferring at lower layers (\ll{pool1} and
\ll{pool2}). Transfer at layers \ll{conv3}, \ll{conv4}, \ll{pool5} works
comparably, but performance deteriorates when moving to further higher layers
(\ll{fc6} and \ll{fc7}). This validates our choice for using an intermediate
layer as a transfer point. We believe the drop in performance at higher layers
is an artifact of the amount of data used for supervision transfer. With a
richer and more diverse dataset of paired images we expect transfer at higher
layers to work similar or better than transfer at mid-layers.

We also conducted some initial experiments with using multiple transfer points.
When transferring at \ll{conv3} and \ll{fc7} we observe performance improves
over transferring at either layer alone, indicating learning is facilitated
when supervision is closer to parameters being learned. We defer exploration of
other choices in this space for future work.

\pparagraph{Is input representation in the form of \hha images still
important?} Given our tool for training {\cnn}s on depth images, we can now
investigate the question whether hand engineering the input representation is
still important. We conduct an experiment in exactly the same settings as
\expt{1D} except that we work with disparity images (replicated to have 3
channels) instead of \hha images. This gives a mAP of 29.2\% as compared to
30.5\% for the \hha images. The difference in performance is smaller than what
\cite{guptaECCV14} reports but still exists, which suggests that encoding
information into geocentric channels through the \hha embedding is still useful
\cite{guptaECCV14}. 

\renewcommand{\arraystretch}{1.4} 
\setlength{\tabcolsep}{5pt}
\begin{table}
\centering
\footnotesize
\resizebox{\linewidth}{!}{
\begin{tabular}{lcccccccccg}
\toprule
& \multicolumn{8}{c}{Train on \mscoco and adapt to \nyu using supervision
transfer} & & \multicolumn{1}{c}{Train on \nyu} \\
\cmidrule(r){2-9} \cmidrule(r){11-11}
               & bed       & chair     & sink      & sofa      & table     & tv        & toilet    & mean      & & mean \\ \midrule
\rgb           & 51.6      & 26.6      & 25.1      & 43.1      & 14.4      & 12.9      & 57.5      & 33.0      & & 35.7 \\
\dd            & 59.4      & 27.1      & 23.8      & 32.2      & 13.0      & 13.6      & 43.8      & 30.4      & & 45.0 \\
\rgb + \dd     & \tb{60.2} & \tb{35.3} & \tb{27.5} & \tb{48.2} & \tb{16.5} & \tb{17.1} & \tb{58.1} & \tb{37.6} & & \tb{54.4}\\
\bottomrule
\end{tabular}}
\caption{\small \textbf{Adapting \rgb object detectors to \rgbd images}:
We transfer object detectors trained on \rgb images (on \mscoco dataset) to
\rgbd images in the \nyu dataset, without using any annotations on depth
images. We do this by learning a model on depth images using supervision
transfer and then use the \rgb object detector trained on the 
representation learned on depth images. We report detection AP(\%) on \nyu
\test set. These transferred detectors work well on depth images even without
using any annotations on depth images. Combining predictions from the \rgb and
depth image improves performance further.}
\tablelabel{zero-shot}
\end{table}

\pparagraph{Applications to zero-shot detection on depth images.} \Supt can be
used to transfer detectors trained on \rgb images to depth images. We do this
by the following steps. We first train detectors on \rgb images. We then split
the network into two parts at an appropriate mid-level point to obtain two
networks $\Gamma^{lower}_{rgb}$ and $\Gamma^{upper}_{rgb}$. We then use the
lower domain specific part of the network $\Gamma^{lower}_{rgb}$ to train a
network $\Gamma^{lower}_{d}$ on depth images to generate the same
representation as the \rgb network $\Gamma^{lower}_{rgb}$. This is done using
the same supervision transfer procedure as before on a set of unlabeled paired
\rgbd images. We then construct a `franken' network with the lower domain
specific part coming from $\Gamma^{lower}_{d}$ and the upper more semantic
network coming from $\Gamma^{upper}_{rgb}$. We then simply use the output of
this franken network on depth images to obtain zero-shot object detection
output.

More specifically, we use \fastrcnn with \alexnet~\cnn to train object
detectors on the \mscoco dataset \cite{mscoco}. We then split the network right
after the convolutional layers \ll{pool5}, and train a network on depth images
to predict the same \ll{pool5} features as this network on unlabeled \rgbd
images from the \nyu dataset (using frames from the \trainval video sequences).
We study all 7 object categories that are shared between \mscoco and \nyu
datasets, and report the performance in \tableref{zero-shot}. We observe our
zero-shot scheme for transferring detectors across modalities works rather
well, and results in good performance. While the \rgb detector trained on
\mscoco obtains a mean \boxAP of 33.0\% on these categories, our zero-shot
detector on \dd images performs comparably and has a mean \boxAP of 30.4\%.
Note that in doing so we have not used any annotations from the \nyu dataset
(\rgb or \dd images). Furthermore, combining predictions from \rgb and \dd
object detectors results in boost over just using the detector on the \rgb
image giving a performance of 37.6\%. Performance when training detectors using
annotations from the \nyu dataset (last column in \tableref{zero-shot}) is
obviously much higher.

\renewcommand{\arraystretch}{1.4} 
\setlength{\tabcolsep}{4pt}
\begin{table}
\centering
\footnotesize
\resizebox{\linewidth}{!}{
\begin{tabular}{lcccc}
\toprule 
method & modality & \rgb Arch. & \dd Arch. & mean \\ \midrule 
\fastrcnn \cite{fastrcnn}                     & \rgb       & \alexnet & -        & 27.8 \\
\fastrcnn \cite{fastrcnn}                     & \rgb       & \vgg     & -        & \tb{38.8} \\
\midrule \midrule
Gupta \etal~\cite{guptaECCV14}                & \rgb + \dd & \alexnet & \alexnet & 38.8 \\
Gupta \etal~\cite{guptaCVPR15a}               & \rgb + \dd & \alexnet & \alexnet & 41.2 \\
Gupta \etal~\cite{guptaECCV14} + \fastrcnn    & \rgb + \dd & \alexnet & \alexnet & 44.4 \\
Our (\textit{supervision transfer})           & \rgb + \dd & \alexnet & \alexnet & \tb{47.1} \\
\midrule
Gupta \etal~\cite{guptaECCV14} + \fastrcnn    & \rgb + \dd & \vgg     & \alexnet & 46.2 \\
Our (\textit{supervision transfer})           & \rgb + \dd & \vgg     & \alexnet & \tb{49.1} \\
\midrule \midrule
Gupta \etal~\cite{guptaECCV14} + \fastrcnn    & \dd        & -        & \alexnet & 34.2 \\
Our (\textit{supervision transfer})           & \dd        & -        & \alexnet & \tb{41.7} \\
\bottomrule 
\end{tabular}}
\caption{\small \textbf{Object detection mean AP(\%) on \nyu \test set:} We
compare our performance against several state-of-the-art methods. \rgb Arch.
and \dd Arch. refers to the \cnn architecture used by the detector. We see when
using just the depth image, our method is able to improve performance from
34.2\% to 41.7\%.  When used in addition to features from the \rgb image, our
learned features improve performance from 44.4\% to 47.1\% (when using
\alexnet~\rgb features) and from 46.2\% to 49.1\% (when using \vgg~\rgb
features) over past methods for learning features from depth images. We see
improvements across almost all categories, performance on individual categories
is tabulated in supplementary material.}
\tablelabel{test}
\end{table}

\pparagraph{Performance on test set.} Finally, we report the performance of our
best performing supervision transfer scheme (\vgg$^* \rightarrow$ \alexnet) on
the \test set in \tableref{test}.  When used with \alexnet for obtaining color
features, we obtain a final performance of 47.1\% which is about 2.7\% higher
than the current state-of-the-art on this task (Gupta \etal \cite{guptaECCV14}
\fastrcnn). We see similar improvements when using \vgg for obtaining color
features (46.2\% to 49.1\%). The improvement when using just the depth image is
much larger, 41.7\% for our final model as compared to 34.2\% for the baseline
model which amounts to a 22\% relative improvement. Note that in obtaining
these performance improvements we are using exactly the same \cnn architecture
and amount of labeled data. We also report performance on the SDS task in
\tableref{sds-test-summary} and obtain state-of-the-art performance of 40.5\% as
compared to previous best 37.5\% \cite{guptaCVPR15a} when using \alexnet, using
\vgg~\cnn for the \rgb image improves performance further to 42.1\%.

\pparagraph{Training Time.} Finally, we report the amount of time it takes to
learn a model using supervision transfer. For \alexnet to \alexnet
supervision transfer we trained for 100K iterations which took a total of 2.5
hours on a NVIDIA k40 GPU. This is a many orders of magnitude faster than
training models from random initialization on ImageNet scale data using class
labels.

\renewcommand{\arraystretch}{1.2} 
\setlength{\tabcolsep}{5pt}
\begin{table}
\centering
\footnotesize
\resizebox{\linewidth}{!}{
\begin{tabular}{lcccccc}
\toprule
& \multicolumn{2}{c}{\rgb}   & \multicolumn{4}{c}{optical flow} \\ \cmidrule(r){2-3}  \cmidrule(r){4-7}
& \cite{gkioxari2015finding} & \cite{gkioxari2015finding} + \cite{fastrcnn}                     & \cite{gkioxari2015finding} & \cite{gkioxari2015finding} + \cite{fastrcnn} & Random Init   & Our \\
&                            &                                                                  & Sup PreTr                 & Sup PreTr                                   & No PreTr & Sup Transfer \\
\midrule
mean AP & 27.0 & \textbf{32.0} & 24.3 & \textbf{38.4} & 31.7 & 35.7 \\
\bottomrule
\end{tabular}}
\caption{\small \textbf{Action Detection AP(\%) on the \jhmdb \test set:} We
report action detection performance on the \test set of \jhmdb using \rgb or
flow images. Right part of the table compares our method \textit{supervision
transfer} against the baseline of random initialization, and the ceiling using
fully supervised pre-training method from \cite{gkioxari2015finding}. Our
method reaches more than half the way towards fully supervised pre-training.}
\tablelabel{jhmdb-test}
\end{table}

\subsection{Transfer to Flow Images}
We now report our experiments for transferring supervision to optical flow
images. We consider the end task of action detection on the \jhmdb dataset. The
task is to detect people doing actions like \ll{catch}, \ll{clap}, \ll{pick},
\ll{run}, \ll{sit} in frames of a video. Performance is measured in terms of
mean average precision as in the standard PASCAL VOC object detection task
and what we used for the \nyu experiments in \secref{nyu}.

A popular technique for getting better performance at such tasks on video data
is to additionally use features computed on the optical flow between the
current frame and the next frame \cite{simonyan2014two, gkioxari2015finding},
and we use our \textit{supervision transfer} scheme to learn features for
optical flow images in this context.

\pparagraph{Detection model} For \jhmdb we use the experimental setup from
Gkioxari and Malik \cite{gkioxari2015finding} and study the 21 class task.
Here again, Gkioxari and Malik build off of \rcnn and we first adapt their
system to use \fastrcnn, and observe similar boosts in performance as for \nyu
when going from \rcnn to \fastrcnn framework (\tableref{jhmdb-test}, full table
with per class performance is in the supplementary material). We denote this
model as \cite{gkioxari2015finding}+\cite{fastrcnn}. We attribute this large
difference in performance to a) bounding box regression and b) number of
iterations used for training.

\pparagraph{\textit{Supervision transfer} performance} We use the videos from
\ucf dataset \cite{UCF101} for our pre-training. Note that we do not use any
labels provided with the \ucf dataset, and simply use the videos as a source of
paired \rgb and flow images. We take 5 frames from each of the $9K$ videos in
the \textit{train1} set. We report performance on \jhmdb \textit{\small test}
set in \tableref{jhmdb-test}. Note that \jhmdb has 3 splits and as in past
work, we report the AP averaged across these 3 splits. 

We report performance for three different schemes for initializing the flow
model: a) \textbf{Random Init} (No PreTr) when the flow network is initialized
randomly using the weight initialization scheme used for training a \rgb model
on ImageNet, b) \textbf{Supervised Pre-training}
(\cite{gkioxari2015finding}+\cite{fastrcnn} Sup PreTr) on flow images from \ucf
for the task of video classification starting from \rgb weights as done by
Gkioxari and Malik \cite{gkioxari2015finding} and c)
\textbf{supervision transfer} (Our Sup Transfer) from an \rgb model to
train optical flow model as per our proposed method. We see that our scheme for
\textit{supervision transfer} improves performance from 31.7\% achieved when
using random initialization to 35.7\%, which is more than half way towards what
fully supervised pre-training can achieve (38.4\%), thereby illustrating the
efficacy of our adaptation scheme.

\pparagraph{Conclusion}
We have presented an algorithm for transfer of learned
representations from a well labeled modality to new unlabeled modalities using
unlabeled paired images from the two modalities. This enables us to learn rich
representations on unlabeled modalities and obtain large boosts in
performance. We believe the advances presented in this paper will allow us to
effectively use new modalities for obtaining better performance on standard
vision tasks. Code, data and pretrained models are available at
\url{https://github.com/s-gupta/fast-rcnn/tree/distillation}.

\paragraph{Acknowledgments: } The authors would like to thank Georgia Gkioxari
for sharing her wisdom and experimental setup for the \ucf and \jhmdb datasets.
This work was supported by ONR SMARTS MURI N00014-09-1-1051, a Berkeley
Graduate Fellowship, a Google Fellowship in Computer Vision and a NSF
Graduate Research Fellowship. We gratefully acknowledge NVIDIA corporation
for the donation of Tesla and Titan GPUs used for this research.

{\small
\bibliographystyle{ieee}
\bibliography{refs-rbg,refs-da}

\begin{thebibliography}{10}\itemsep=-1pt

\bibitem{aytar-iccv11}
Y.~Aytar and A.~Zisserman.
\newblock Tabula rasa: Model transfer for object category detection.
\newblock In {\em ICCV}, 2011.

\bibitem{bucilua2006model}
C.~Bucilua, R.~Caruana, and A.~Niculescu-Mizil.
\newblock Model compression.
\newblock In {\em ACM SIGKDD}, 2006.

\bibitem{chatfield2014return}
K.~Chatfield, K.~Simonyan, A.~Vedaldi, and A.~Zisserman.
\newblock Return of the devil in the details: Delving deep into convolutional
  nets.
\newblock In {\em BMVC}, 2014.

\bibitem{ref:dlid}
S.~Chopra, S.~Balakrishnan, and R.~Gopalan.
\newblock {DLID}: Deep learning for domain adaptation by interpolating between
  domains.
\newblock In {\em ICML Workshop on Challenges in Representation Learning},
  2013.

\bibitem{christoudias2010learning}
C.~M. Christoudias, R.~Urtasun, M.~Salzmann, and T.~Darrell.
\newblock Learning to recognize objects from unseen modalities.
\newblock In {\em Computer Vision--ECCV 2010}, pages 677--691. Springer, 2010.

\bibitem{imagenet_cvpr09}
J.~Deng, W.~Dong, R.~Socher, L.-J. Li, K.~Li, and L.~Fei-Fei.
\newblock Image{N}et: A large-scale hierarchical image database.
\newblock In {\em CVPR}, 2009.

\bibitem{donahueDecaf}
J.~Donahue, Y.~Jia, O.~Vinyals, J.~Hoffman, N.~Zhang, E.~Tzeng, and T.~Darrell.
\newblock Decaf: A deep convolutional activation feature for generic visual
  recognition.
\newblock In {\em ICML}, 2014.

\bibitem{duan-icml12}
L.~Duan, D.~Xu, and I.~W. Tsang.
\newblock Learning with augmented features for heterogeneous domain adaptation.
\newblock In {\em ICML}, 2012.

\bibitem{frome2013devise}
A.~Frome, G.~S. Corrado, J.~Shlens, S.~Bengio, J.~Dean, T.~Mikolov, et~al.
\newblock Devise: A deep visual-semantic embedding model.
\newblock In {\em NIPS}, 2013.

\bibitem{ganin2015}
Y.~{Ganin} and V.~{Lempitsky}.
\newblock {Unsupervised Domain Adaptation by Backpropagation}.
\newblock {\em ArXiv e-prints}, Sept. 2014.

\bibitem{da-mmd}
M.~Ghifary, W.~B. Kleijn, and M.~Zhang.
\newblock Domain adaptive neural networks for object recognition.
\newblock {\em CoRR}, abs/1409.6041, 2014.

\bibitem{fastrcnn}
R.~Girshick.
\newblock Fast {R-CNN}.
\newblock In {\em ICCV}, 2015.

\bibitem{girshickCVPR14}
R.~Girshick, J.~Donahue, T.~Darrell, and J.~Malik.
\newblock Rich feature hierarchies for accurate object detection and semantic
  segmentation.
\newblock In {\em CVPR}, 2014.

\bibitem{gkioxari2015finding}
G.~Gkioxari and J.~Malik.
\newblock Finding action tubes.
\newblock In {\em CVPR}, 2015.

\bibitem{gong-cvpr12}
B.~Gong, Y.~Shi, F.~Sha, and K.~Grauman.
\newblock Geodesic flow kernel for unsupervised domain adaptation.
\newblock In {\em CVPR}, 2012.

\bibitem{guptaCVPR15a}
S.~Gupta, P.~Arbel{\'{a}}ez, R.~Girshick, and J.~Malik.
\newblock Aligning {3D} models to {RGB-D} images of cluttered scenes.
\newblock In {\em CVPR}, 2015.

\bibitem{guptaECCV14}
S.~Gupta, R.~Girshick, P.~Arbel{\'a}ez, and J.~Malik.
\newblock Learning rich features from {RGB-D} images for object detection and
  segmentation.
\newblock In {\em ECCV}, 2014.

\bibitem{Hariharan:EECS-2015-193}
B.~Hariharan.
\newblock {\em Beyond Bounding Boxes: Precise Localization of Objects in
  Images}.
\newblock PhD thesis, EECS Department, University of California, Berkeley, Aug
  2015.

\bibitem{hariharanECCV14}
B.~Hariharan, P.~Arbel{\'a}ez, R.~Girshick, and J.~Malik.
\newblock Simultaneous detection and segmentation.
\newblock In {\em ECCV}, 2014.

\bibitem{hariharan2015hypercolumns}
B.~Hariharan, P.~Arbel\'{a}ez, R.~Girshick, and J.~Malik.
\newblock Hypercolumns for object segmentation and fine-grained localization.
\newblock In {\em CVPR}, 2015.

\bibitem{kaiming2014spatial}
K.~He, X.~Zhang, S.~Ren, and J.~Sun.
\newblock Spatial pyramid pooling in deep convolutional networks for visual
  recognition.
\newblock In {\em ECCV}, 2014.

\bibitem{hinton2014distilling}
G.~E. Hinton, O.~Vinyals, and J.~Dean.
\newblock Distilling the knowledge in a neural network.
\newblock In {\em NIPS 2014 Deep Learning Workshop}, 2014.

\bibitem{lsda}
J.~Hoffman, S.~Guadarrama, E.~Tzeng, R.~Hu, J.~Donahue, R.~Girshick,
  T.~Darrell, and K.~Saenko.
\newblock {LSDA}: Large scale detection through adaptation.
\newblock In {\em NIPS}, 2014.

\bibitem{hoffman-iclr13}
J.~Hoffman, E.~Rodner, J.~Donahue, K.~Saenko, and T.~Darrell.
\newblock Efficient learning of domain-invariant image representations.
\newblock In {\em ICLR}, 2013.

\bibitem{hoffman-iclr14}
J.~Hoffman, E.~Tzeng, J.~Donahue, , Y.~Jia, K.~Saenko, and T.~Darrell.
\newblock One-shot learning of supervised deep convolutional models.
\newblock In {\em arXiv 1312.6204; presented at ICLR Workshop}, 2014.

\bibitem{b3do}
A.~Janoch, S.~Karayev, Y.~Jia, J.~T. Barron, M.~Fritz, K.~Saenko, and
  T.~Darrell.
\newblock A category-level {3D} object dataset: Putting the kinect to work.
\newblock In {\em Consumer Depth Cameras for Computer Vision}. 2013.

\bibitem{Jhuang:ICCV:2013}
H.~Jhuang, J.~Gall, S.~Zuffi, C.~Schmid, and M.~J. Black.
\newblock Towards understanding action recognition.
\newblock In {\em ICCV}, 2013.

\bibitem{jiaCaffe}
Y.~Jia.
\newblock {Caffe}: An open source convolutional architecture for fast feature
  embedding.
\newblock \url{http://caffe.berkeleyvision.org/}, 2013.

\bibitem{jones1992determining}
D.~G. Jones and J.~Malik.
\newblock Determining three-dimensional shape from orientation and spatial
  frequency disparities.
\newblock In {\em ECCV}, 1992.

\bibitem{koenderink1987representation}
J.~J. Koenderink and A.~J. van Doorn.
\newblock Representation of local geometry in the visual system.
\newblock {\em Biological cybernetics}, 55(6):367--375, 1987.

\bibitem{krizhevsky2012imagenet}
A.~Krizhevsky, I.~Sutskever, and G.~Hinton.
\newblock Image{N}et classification with deep convolutional neural networks.
\newblock In {\em NIPS}, 2012.

\bibitem{kulis-cvpr11}
B.~Kulis, K.~Saenko, and T.~Darrell.
\newblock What you saw is not what you get: Domain adaptation using asymmetric
  kernel transforms.
\newblock In {\em CVPR}, 2011.

\bibitem{lecun-89e}
Y.~LeCun, B.~Boser, J.~S. Denker, D.~Henderson, R.~E. Howard, W.~Hubbard, and
  L.~D. Jackel.
\newblock Backpropagation applied to handwritten zip code recognition.
\newblock {\em Neural Computation}, 1989.

\bibitem{lenc15understanding}
K.~Lenc and A.~Vedaldi.
\newblock Understanding image representations by measuring their equivariance
  and equivalence.
\newblock In {\em CVPR}, 2015.

\bibitem{mscoco}
T.~Lin, M.~Maire, S.~Belongie, J.~Hays, P.~Perona, D.~Ramanan, P.~Doll{\'a}r,
  and C.~L. Zitnick.
\newblock Microsoft {COCO}: Common objects in context.
\newblock In {\em ECCV}, 2014.

\bibitem{long_shelhamer_fcn}
J.~Long, E.~Shelhamer, and T.~Darrell.
\newblock Fully convolutional networks for semantic segmentation.
\newblock In {\em CVPR}, 2015.

\bibitem{long2015dan}
M.~Long and J.~Wang.
\newblock Learning transferable features with deep adaptation networks.
\newblock {\em CoRR}, abs/1502.02791, 2015.

\bibitem{ngiam2011multimodal}
J.~Ngiam, A.~Khosla, M.~Kim, J.~Nam, H.~Lee, and A.~Y. Ng.
\newblock Multimodal deep learning.
\newblock In {\em Proceedings of the 28th International Conference on Machine
  Learning (ICML-11)}, pages 689--696, 2011.

\bibitem{romero2014fitnets}
A.~Romero, N.~Ballas, S.~E. Kahou, A.~Chassang, C.~Gatta, and Y.~Bengio.
\newblock Fitnets: Hints for thin deep nets.
\newblock {\em arXiv preprint arXiv:1412.6550}, 2014.

\bibitem{overfeat}
P.~Sermanet, D.~Eigen, X.~Zhang, M.~Mathieu, R.~Fergus, and Y.~LeCun.
\newblock Overfeat: Integrated recognition, localization and detection using
  convolutional networks.
\newblock In {\em ICLR}, 2014.

\bibitem{silbermanECCV12}
N.~Silberman, D.~Hoiem, P.~Kohli, and R.~Fergus.
\newblock Indoor segmentation and support inference from {RGBD} images.
\newblock In {\em ECCV}, 2012.

\bibitem{silbermanECCV14}
N.~Silberman, D.~Sontag, and R.~Fergus.
\newblock Instance segmentation of indoor scenes using a coverage loss.
\newblock In {\em ECCV}, 2014.

\bibitem{simonyan2014two}
K.~Simonyan and A.~Zisserman.
\newblock Two-stream convolutional networks for action recognition in videos.
\newblock In {\em NIPS}, 2014.

\bibitem{simonyan2014very}
K.~Simonyan and A.~Zisserman.
\newblock Very deep convolutional networks for large-scale image recognition.
\newblock {\em arXiv preprint arXiv:1409.1556}, 2014.

\bibitem{socher2013zero}
R.~Socher, M.~Ganjoo, C.~D. Manning, and A.~Ng.
\newblock Zero-shot learning through cross-modal transfer.
\newblock In {\em NIPS}, 2013.

\bibitem{song2015sun}
S.~Song, S.~P. Lichtenberg, and J.~Xiao.
\newblock Sun rgb-d: A rgb-d scene understanding benchmark suite.
\newblock In {\em CVPR}, 2015.

\bibitem{UCF101}
K.~Soomro, A.~R. Zamir, and M.~Shah.
\newblock Ucf101: A dataset of 101 human action classes from videos in the
  wild.
\newblock In {\em CRCV-TR-12-01}, 2012.

\bibitem{JMLR:v15:srivastava14b}
N.~Srivastava and R.~Salakhutdinov.
\newblock Multimodal learning with deep boltzmann machines.
\newblock {\em JMRL}, 2014.

\bibitem{tulsianiCVPR15}
S.~Tulsiani and J.~Malik.
\newblock Viewpoints and keypoints.
\newblock In {\em CVPR}, 2015.

\bibitem{ddc}
E.~Tzeng, J.~Hoffman, N.~Zhang, K.~Saenko, and T.~Darrell.
\newblock Deep domain confusion: Maximizing for domain invariance.
\newblock {\em CoRR}, abs/1412.3474, 2014.

\bibitem{weber1995robust}
J.~Weber and J.~Malik.
\newblock Robust computation of optical flow in a multi-scale differential
  framework.
\newblock {\em IJCV}, 1995.

\end{thebibliography}
}
\balance
\clearpage

\setcounter{section}{0}
\renewcommand\thesection{\Alph{section}}

\section{Supplementary Material}
\begin{enumerate}
\item \textbf{Per category average precision:} We report per category numbers
for summary tables on \test sets in the main paper.  
\item \textbf{Sample Detection and SDS output: } We show sample detections and
SDS output for the categories we study. We sample 18 detections uniformly from
the top $k$ ($= 0.75 \times$ number of instances) detections for each category:
bed (\figref{bed}), chair (\figref{chair}), sofa (\figref{sofa}), toilet
(\figref{toilet}), table (\figref{table}). 
\end{enumerate}
\insertFIG{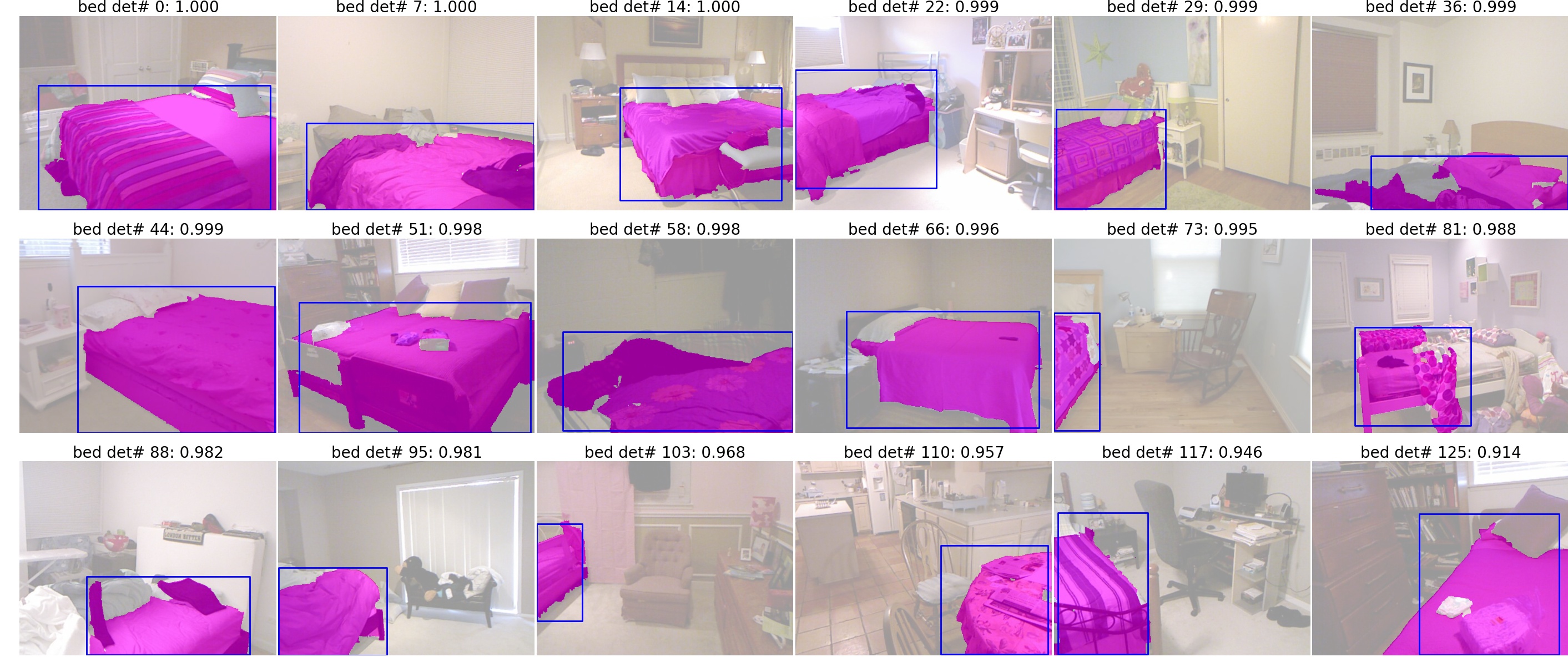}{0.8}
\insertFIG{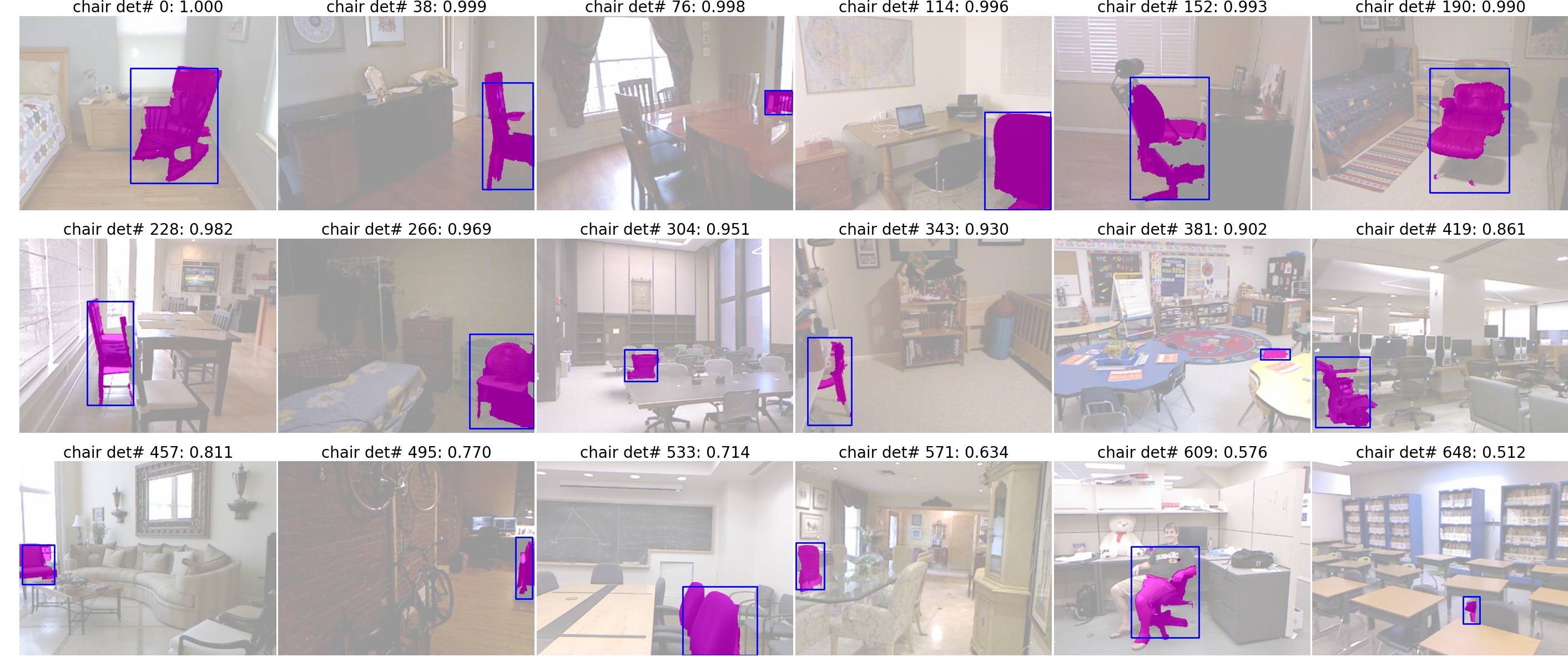}{0.8}
\insertFIG{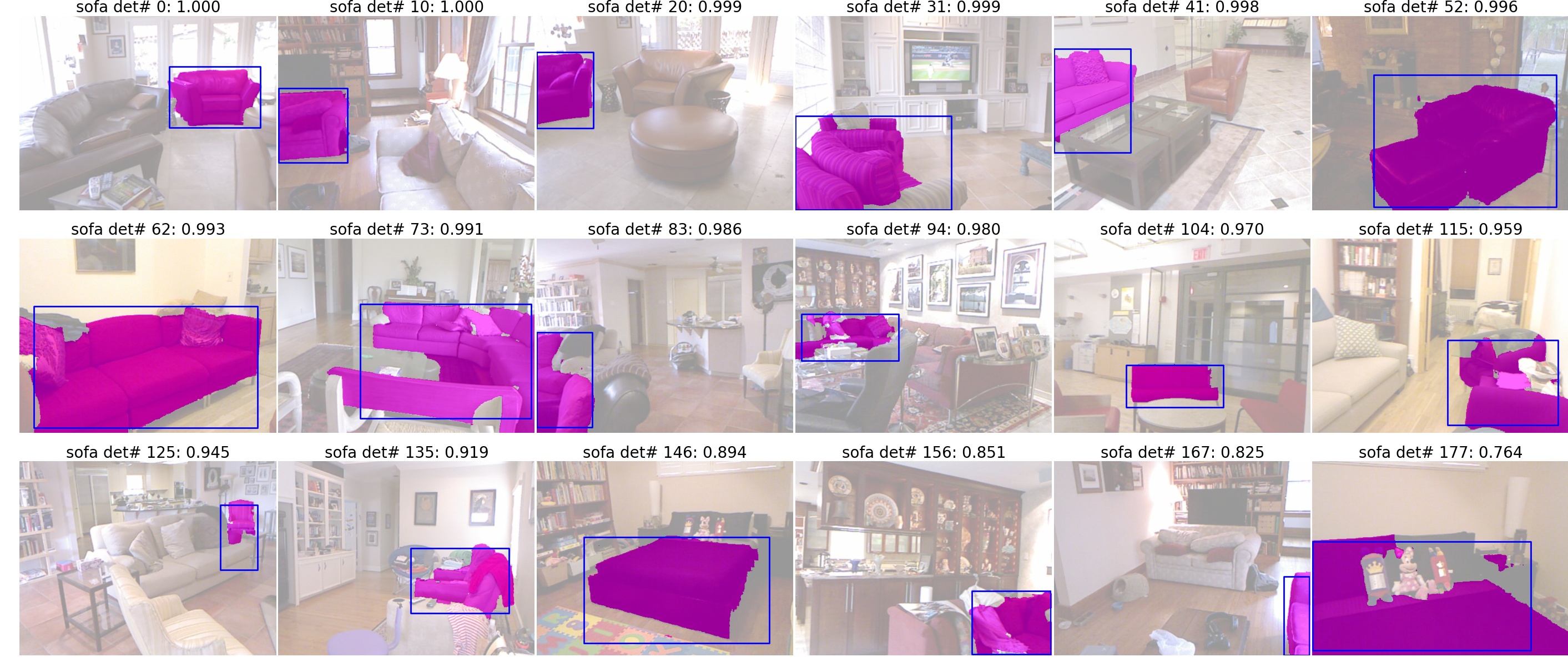}{0.8}
\insertFIG{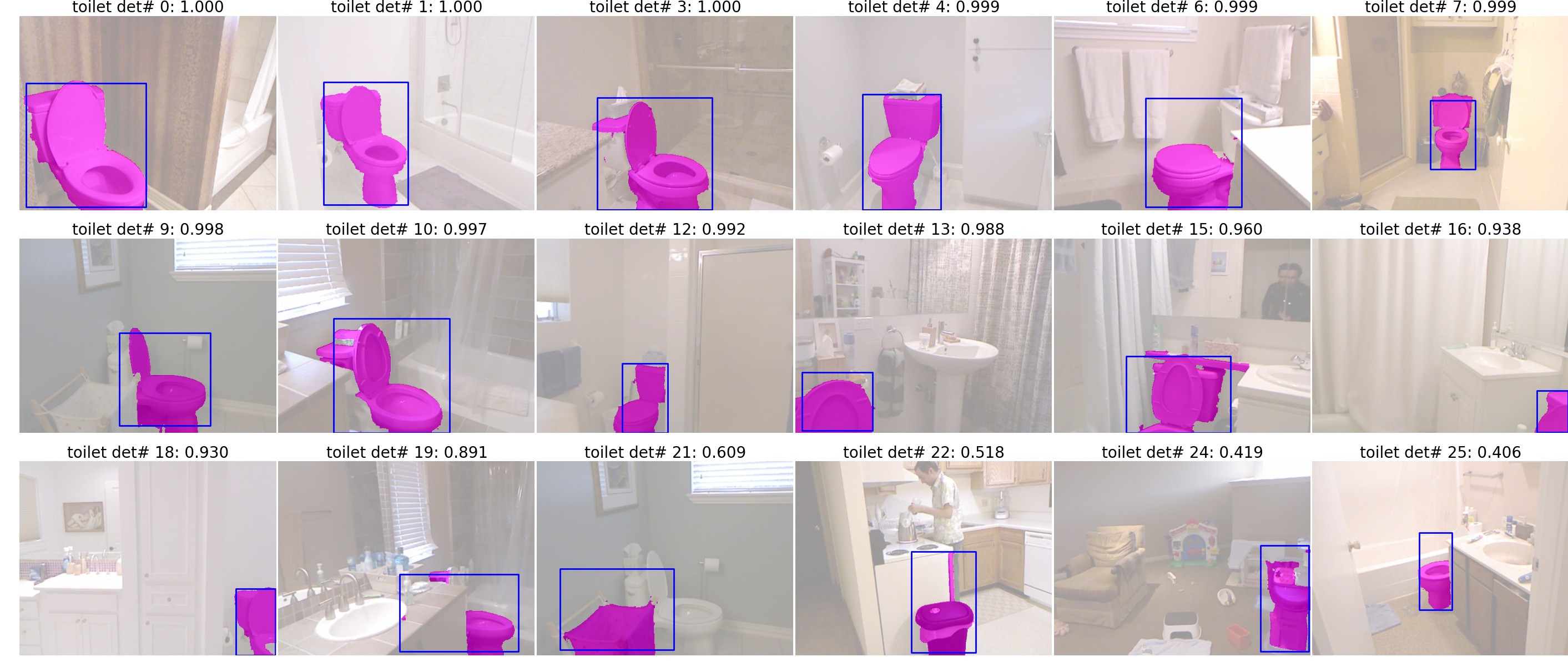}{0.8}
\insertFIG{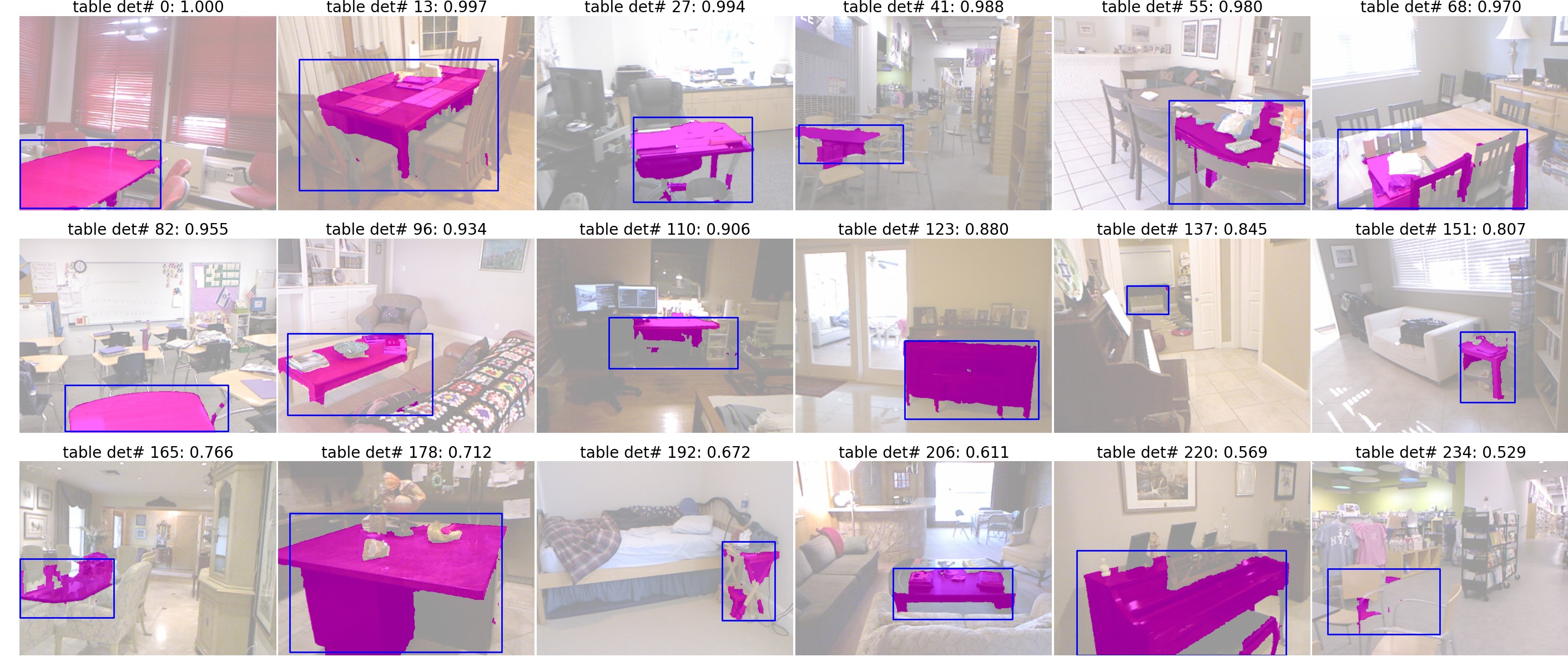}{0.8}
\renewcommand{\arraystretch}{1.4} 
\setlength{\tabcolsep}{6pt}
\begin{table*}
\centering
\footnotesize
\resizebox{1.0\linewidth}{!}{
\begin{tabular}{lcccccccccccccccccccccccgg}
\toprule 
\verticala{method} & \verticala{modality} & \verticala{\rgb Arch.} &
\verticala{\dd Arch.} & \verticala{metric} & \verticala{bath tub} &
\verticala{bed} & \verticala{book shelf} & \verticala{box} & \verticala{chair}
& \verticala{counter} & \verticala{desk} & \verticala{door} &
\verticala{dresser} & \verticala{garbage bin} & \verticala{lamp} &
\verticala{monitor} & \verticala{night stand} & \verticala{pillow} &
\verticala{sink} & \verticala{sofa} & \verticala{table} & \verticala{tele
vision} & \verticala{toilet} & \verticala{mean} \\ \midrule
\cite{hariharan2015hypercolumns}    & \rgb       & \alexnet & -        & \regionAP 0.5 & 8.9  & 45.2 & 12.6 & 1.4 & 20.6 & 24.3 & 4.2  & 19.3 & 27.2 & 20.1 & 26.2 & 39.2 & 21.3 & 23.7 & 27.6 & 25.2 & 8.2  & 35.3 & 54.3 & 23.4  \\
Gupta \etal~\cite{guptaCVPR15a}     & \rgb + \dd & \alexnet & \alexnet & \regionAP 0.5 & 42.0 & 65.1 & 12.7 & 5.1 & 42.0 & 42.1 & 9.5  & 20.5 & 38.0 & 50.3 & 32.8 & 54.5 & 38.2 & 42.0 & 39.4 & 46.6 & 14.8 & 48.0 & 68.4 & 37.5  \\
Our (\textit{supervision transfer}) & \rgb + \dd & \alexnet & \alexnet & \regionAP 0.5 & 31.5 & 68.7 & 22.3 & 4.0 & 39.6 & 43.3 & 11.2 & 25.1 & 52.1 & 42.5 & 45.0 & 61.8 & 47.5 & 41.3 & 48.5 & 49.7 & 18.1 & 49.5 & 68.4 & 40.5  \\
\cite{hariharan2015hypercolumns}    & \rgb       & \vgg     & -        & \regionAP 0.5 & 17.5 & 59.5 & 10.7 & 2.6 & 32.7 & 28.3 & 5.8  & 22.4 & 42.2 & 32.9 & 34.3 & 54.5 & 26.7 & 28.9 & 36.1 & 38.3 & 9.6  & 44.1 & 62.5 & 31.0  \\
Our (\textit{supervision transfer}) & \rgb + \dd & \vgg     & \alexnet & \regionAP 0.5 & 42.2 & 69.5 & 18.3 & 5.3 & 45.1 & 41.9 & 10.7 & 29.1 & 55.2 & 48.1 & 45.1 & 62.3 & 46.9 & 42.0 & 46.0 & 54.8 & 17.2 & 49.0 & 71.1 & 42.1  \\
\midrule
\cite{hariharan2015hypercolumns}    & \rgb       & \alexnet & -        & \regionAP 0.7 & 5.9  & 28.6 & 2.8 & 0.6 & 6.6  & 6.0  & 1.2 & 9.6  & 16.8 & 15.8 & 8.4  & 16.1 & 17.1 & 15.5 & 11.9 & 12.9 & 1.8 & 33.7 & 44.0 & 13.4 \\
Gupta \etal~\cite{guptaCVPR15a}     & \rgb + \dd & \alexnet & \alexnet & \regionAP 0.7 & 13.8 & 46.0 & 2.4 & 3.0 & 17.3 & 15.0 & 2.6 & 9.9  & 25.8 & 45.4 & 6.9  & 37.5 & 24.3 & 25.5 & 19.6 & 27.9 & 7.6 & 44.9 & 38.7 & 21.8 \\
Our (\textit{supervision transfer}) & \rgb + \dd & \alexnet & \alexnet & \regionAP 0.7 & 13.3 & 50.6 & 5.3 & 1.3 & 15.9 & 14.2 & 2.6 & 15.6 & 50.0 & 34.0 & 14.0 & 36.4 & 33.8 & 26.3 & 20.8 & 27.7 & 6.9 & 44.9 & 68.4 & 25.4 \\
\cite{hariharan2015hypercolumns}    & \rgb       & \vgg     & -        & \regionAP 0.7 & 6.6  & 35.7 & 0.4 & 1.6 & 9.4  & 7.2  & 1.1 & 16.5 & 29.3 & 29.1 & 11.3 & 33.3 & 19.5 & 19.9 & 17.2 & 17.9 & 1.7 & 35.7 & 43.4 & 17.7 \\
Our (\textit{supervision transfer}) & \rgb + \dd & \vgg     & \alexnet & \regionAP 0.7 & 13.0 & 56.1 & 6.9 & 2.5 & 17.9 & 14.8 & 4.3 & 18.6 & 51.7 & 36.2 & 16.2 & 42.2 & 32.3 & 26.9 & 20.4 & 32.5 & 6.3 & 44.4 & 68.7 & 26.9 \\
\bottomrule
\end{tabular}}
\caption{\small \textbf{Region Detection \regionAP(\%) on \nyu \test set:} We
report per class \regionAP for the SDS experiments in \tableref{sds-test-summary} in the main paper.}
\tablelabel{sds-test_supp}
\end{table*}

\renewcommand{\arraystretch}{1.4} 
\setlength{\tabcolsep}{3.5pt}
\begin{table*}
\centering
\footnotesize
\resizebox{\linewidth}{!}{
\begin{tabular}{lccccccccccccccccccccccg}
\toprule 
method & modality & \rgb Arch. & \dd Arch. & \verticala{bath tub} & \verticala{bed} & \verticala{book shelf}
& \verticala{box} & \verticala{chair} & \verticala{counter} & \verticala{desk} &
\verticala{door} & \verticala{dresser} & \verticala{garbage bin} & \verticala{lamp}
& \verticala{monitor} & \verticala{night stand} & \verticala{pillow} &
\verticala{sink} & \verticala{sofa} & \verticala{table} & \verticala{tele vision} &
\verticala{toilet} & \verticala{mean} \\ \midrule
\fastrcnn \cite{fastrcnn}                     & \rgb       & \alexnet & -        & 7.9       & 51.2      & 37.0      & 1.5       & 31.3      & 35.4      & 9.4       & 22.4      & 28.9      & 19.3      & 31.0      & 35.9      & 24.1      & 26.4      & 24.6      & 39.7      & 16.6      & 32.9      & 53.5      & 27.8 \\
\fastrcnn \cite{fastrcnn}                     & \rgb       & \vgg     & -        & \tb{37.4} & \tb{69.1} & \tb{47.0} & \tb{ 2.9} & \tb{44.4} & \tb{48.6} & \tb{11.5} & \tb{28.7} & \tb{43.1} & \tb{33.6} & \tb{32.9} & \tb{50.9} & \tb{32.6} & \tb{34.4} & \tb{39.0} & \tb{50.3} & \tb{24.5} & \tb{44.1} & \tb{61.5} & \tb{38.8} \\
\midrule \midrule
Gupta \etal~\cite{guptaECCV14}                & \rgb + \dd & \alexnet & \alexnet & 36.4      & 70.8      & 35.1      & 3.6       & 47.3      & 46.8      & 14.9      & 23.3      & 38.6      & 43.9      & 37.6      & 52.7      & 40.7      & 42.4      & 43.5      & 51.6      & 22.0      & 38.0      & 47.7      & 38.8 \\
Gupta \etal~\cite{guptaCVPR15a}               & \rgb + \dd & \alexnet & \alexnet & 39.4      & 73.6      & 38.4      & \tb{ 5.9} & 50.1      & 47.3      & 14.6      & 24.4      & 42.9      & \tb{51.5} & 36.2      & 52.1      & 41.5      & 42.9      & 42.6      & 54.6      & 25.4      & 48.6      & 50.2      & 41.2 \\
Gupta \etal~\cite{guptaECCV14} + \fastrcnn    & \rgb + \dd & \alexnet & \alexnet & 37.1      & 78.3      & \tb{48.5} & 3.3       & 45.3      & 54.6      & \tb{21.9} & 28.5      & 48.6      & 41.9      & 42.5      & 60.6      & 49.2      & 43.7      & 40.2      & 62.1      & 29.2      & 44.3      & \tb{63.6} & 44.4 \\
Our (\textit{supervision transfer})           & \rgb + \dd & \alexnet & \alexnet & \tb{45.6} & \tb{78.7} & \tb{48.5} & 4.3       & \tb{50.5} & \tb{57.8} & 21.4      & \tb{29.6} & \tb{54.0} & 41.6      & \tb{45.4} & \tb{61.2} & \tb{57.9} & \tb{47.3} & \tb{48.9} & \tb{63.2} & \tb{29.5} & \tb{50.0} & 60.1      & \tb{47.1} \\
\midrule
Gupta \etal~\cite{guptaECCV14} + \fastrcnn    & \rgb + \dd & \vgg     & \alexnet & 47.2      & 80.4      & \tb{52.8} & 4.2       & 49.7      & 53.0      & \tb{22.4} & 33.7      & 52.1      & 44.4      & 39.2      & \tb{64.6} & 47.5      & 45.1      & 42.1      & 63.2      & 31.4      & 42.1      & 63.0      & 46.2 \\
Our (\textit{supervision transfer})           & \rgb + \dd & \vgg     & \alexnet & \tb{50.6} & \tb{81.0} & 52.6      & \tb{ 5.4} & \tb{53.0} & \tb{56.1} & 20.9      & \tb{34.6} & \tb{57.9} & \tb{46.2} & \tb{42.5} & 62.9      & \tb{54.7} & \tb{49.1} & \tb{50.0} & \tb{65.9} & \tb{31.9} & \tb{50.1} & \tb{68.0} & \tb{49.1} \\
\midrule \midrule
Gupta \etal~\cite{guptaECCV14} + \fastrcnn    & \dd        & -        & \alexnet & 28.8      & 79.1      & 30.3      & 1.5       & 42.6      & 42.7      & \tb{17.2} & 13.4      & 31.6      & 23.7      & 29.9      & 40.2      & 36.2      & 40.5      & 23.4      & 59.9      & 26.4      & 24.9      & 58.3      & 34.2 \\
Our (\textit{supervision transfer})           & \dd        & -        & \alexnet & \tb{31.2} & \tb{80.7} & \tb{38.6} & \tb{ 2.5} & \tb{52.2} & \tb{52.2} & \tb{17.2} & \tb{18.2} & \tb{50.8} & \tb{35.1} & \tb{37.4} & \tb{51.3} & \tb{50.5} & \tb{43.4} & \tb{41.0} & \tb{63.5} & \tb{29.3} & \tb{37.4} & \tb{59.8} & \tb{41.7} \\
\bottomrule
\end{tabular}}
\caption{\small \textbf{Object Detection AP(\%) on \nyu \test set:} We compare
our performance against several state-of-the-art methods. \rgb Arch. and \dd
Arch. refers to the \cnn architecture used by the detector. We see when using
just the depth image, our method is able to improve performance from 34.2\% to
41.7\%.  When used in addition to features from the \rgb image, our learned
features improve performance from 44.4\% to 47.1\% (when using \alexnet~\rgb
features) and from 46.2\% to 49.1\% (when using \vgg~\rgb features) over past
methods for learning features from depth images. Analogous to summary
\tableref{test} in the main paper.}
\tablelabel{test_supp}
\end{table*}

\renewcommand{\arraystretch}{1.5} 
\setlength{\tabcolsep}{3.5pt}
\begin{table*}
\centering
\footnotesize
\resizebox{\linewidth}{!}{
\begin{tabular}{lcc  cccccccccccccccccccccg}
\toprule method & Supervision & modality &
\vertical{brush hair} & \vertical{catch} & \vertical{clap} & \vertical{climb
stairs} & \vertical{golf} & \vertical{jump} & \vertical{kick ball} &
\vertical{pick} & \vertical{pour} & \vertical{pullup} & \vertical{push} &
\vertical{run} & \vertical{shoot ball} & \vertical{shoot bow} & \vertical{shoot
gun} & \vertical{sit} & \vertical{stand} & \vertical{swing baseball} &
\vertical{throw} & \vertical{walk} & \vertical{wave} & \vertical{mean} \\ \midrule
 Gkioxari \etal~\cite{gkioxari2015finding}           &                 & \rgb      & \tb{55.8} & 25.5      & 25.1      & \tb{24.0} & 77.5      & 1.9       & 5.3       & \tb{21.4} & 68.6      & 71.0      & 15.4      & 6.3       & 4.6       & 41.1      & 28.0      & 9.4       & 8.2       & 19.9      & 17.8      & \tb{29.2} & 11.5         & 27.0 \\
 Gkioxari \etal~\cite{gkioxari2015finding}+\fastrcnn &                 & \rgb      & 47.2      & \tb{35.2} & \tb{30.1} & 23.9      & \tb{84.4} & \tb{ 2.2} & \tb{10.6} & 20.7      & \tb{79.7} & \tb{78.7} & \tb{25.2} & \tb{14.4} & \tb{ 8.7} & \tb{45.3} & \tb{34.2} & \tb{11.7} & \tb{13.3} & \tb{39.0} & \tb{19.1} & 23.9      & \tb{23.9}    & \tb{32.0} \\
 \midrule \midrule
 Gkioxari \etal~\cite{gkioxari2015finding}           & Sup. PreTr.   & flow & 32.3      & 5.0       & 35.6      & 30.1      & 58.0      & 7.8       & 2.6       & 16.4      & 55.0      & 72.3      & 8.5       & 6.1       & 3.9       & 47.8      & 7.3       & 24.9      & 26.3      & 36.3      & 4.5       & 22.1      & 7.6       & 24.3 \\
 Gkioxari \etal~\cite{gkioxari2015finding}+\fastrcnn & Sup. PreTr.   & flow & \tb{54.9} & 17.0      & \tb{52.5} & \tb{56.5} & \tb{81.2} & \tb{15.0} & \tb{10.9} & \tb{28.9} & \tb{72.7} & \tb{86.6} & \tb{20.4} & \tb{17.5} & \tb{10.2} & \tb{61.9} & 25.5      & \tb{31.4} & \tb{42.4} & \tb{53.8} & 10.9      & \tb{38.6} & \tb{17.3} & \tb{38.4} \\
 no init                                             & No PreTr.     & flow & 44.3      & 11.0      & 42.8      & 38.7      & 76.1      & 10.6      & 6.6       & 23.1      & 62.1      & 84.0      & 15.4      & 9.6       & 6.8       & 60.0      & 22.8      & 29.6      & 26.8      & 43.5      & 10.7      & 30.8      & 9.8       & 31.7 \\
 Our (\textit{supervision transfer})                 & Sup. Transfer & flow & 54.6      & \tb{17.7} & 45.1      & 54.9      & 80.3      & 14.6      & 9.7       & 28.2      & 69.3      & 84.8      & 19.9      & 15.6      & 7.2       & 49.6      & \tb{29.4} & 29.5      & 28.4      & 49.5      & \tb{11.6} & 36.3      & 13.0      & 35.7 \\
 \bottomrule
\end{tabular}}
\caption{\small \textbf{Action Detection AP(\%) on the \jhmdb \test set:} We
report action detection performance on the \test set of \jhmdb using \rgb or
flow images. Bottom part of the table, compares our method \textit{supervision
transfer} against the baseline of random initialization, and the ceiling using
fully supervised pre-training method from \cite{gkioxari2015finding}. Our
method reaches more than half the way towards fully supervised pre-training.
Analogous to \tableref{jhmdb-test} in the main paper.} 
\tablelabel{jhmdb-test_supp}
\end{table*}

\section{Document Changelog}
\paragraph{v1} Initial version

\paragraph{v2} Major changes: additional discussion of multi-modal literature,
visualization of neural activations in \figref{filters}(g-i), additional
experiments about quality of intermediate layers, performance as a function of
transfer point, utility of \hha embedding over disparity images, zero-shot
detection on depth images. Minor edits all over the text. 
\balance

\end{document}